\documentclass[runningheads]{llncs}

\usepackage{eccv}

\titlerunning{CroMo-Mixup}
\usepackage{multirow} 

\usepackage{eccvabbrv}

\usepackage[pdftex]{graphicx}
\usepackage{booktabs}
\usepackage{algorithm}
\usepackage{wrapfig}
\usepackage{lipsum}
\usepackage[accsupp]{axessibility}  
\usepackage{amsmath}
\usepackage{changepage}
\DeclareMathOperator*{\argmin}{arg\,min} 

\usepackage{hyperref}

\usepackage{orcidlink}
\usepackage{pifont}

\begin{document}
\newcommand*\samethanks[1][\value{footnote}]{\footnotemark[#1]}
\title{CroMo-Mixup: Augmenting Cross-Model Representations for Continual Self-Supervised Learning} 

\titlerunning{CroMo-Mixup}

\author{Erum Mushtaq\inst{1}\orcidlink{0009-0008-8151-7920} \and
Duygu Nur Yaldiz\inst{1}\orcidlink{0009-0008-1340-5978} \and
Yavuz Faruk Bakman\inst{1}\orcidlink{0009-0003-3655-7943} \and
Jie Ding \inst{2}\orcidlink{0000-0002-3584-6140} \and
Chenyang Tao \inst{3}\orcidlink{0000-0002-2155-8180}\thanks{This work does not relate to their position at Amazon.} \and
Dimitrios Dimitriadis \inst{3}\orcidlink{0000-0001-8483-0105}\samethanks \and
Salman Avestimehr \inst{1}\orcidlink{0000-0003-3102-0867} 
}
%
\authorrunning{E.~Mushtaq et al.}

\institute{University of Southern California\\
\email{\{emushtaq, yaldiz, ybakman, avestime\}@usc.edu} \and
University of Minnesota, 
\email{dingj@umn.edu} \and
Amazon AI, 
\email{\{chenyt, dbdim\}@amazon.com}}

\maketitle

\begin{abstract}
Continual self-supervised learning (CSSL) learns a series of tasks sequentially on the unlabeled data. Two main challenges of continual learning are catastrophic forgetting and task confusion. While CSSL problem has been studied to address the catastrophic forgetting challenge, little work has been done to address the task confusion aspect. In this work, we show through extensive experiments that self-supervised learning (SSL) can make CSSL more susceptible to the task confusion problem, particularly in less diverse settings of class incremental learning because different classes belonging to different tasks are not trained concurrently. Motivated by this challenge, we present a novel cross-model feature Mixup (CroMo-Mixup) framework that addresses this issue through two key components: 1) Cross-Task data Mixup, which mixes samples across tasks to enhance negative sample diversity; and 2) Cross-Model feature Mixup, which learns similarities between embeddings obtained from current and old models of the mixed sample and the original images, facilitating cross-task class contrast learning and old knowledge retrieval. We evaluate the effectiveness of CroMo-Mixup to improve both Task-ID prediction and average linear accuracy across all tasks on three datasets, CIFAR10, CIFAR100, and tinyImageNet under different class-incremental learning settings. We validate the compatibility of CroMo-Mixup on four state-of-the-art SSL objectives. 
Code is available at \url{https://github.com/ErumMushtaq/CroMo-Mixup}.

\end{abstract}

\section{Introduction}
Self-supervised learning (SSL) has advanced significantly in recent years, demonstrating performance on par with supervised learning on diverse computer vision tasks, including image classification \cite{infomax}, segmentation \cite{zeng2019sese}, and object detection \cite{byol}. However, many existing SSL works assume the availability of large, unbiased datasets for model training, which may not always represent a realistic scenario. Data often becomes available progressively in many real-world applications such as self-driving cars \cite{zhang2022claire} and conversational agents \cite{li2020compositional, lin2023speciality}. Given the sequential nature of the data generation process of these real-world applications, it can be impractical to obtain human annotations on-the-fly. Therefore, the exploration of SSL for continual learning holds significant importance.
\begin{figure}[t!]
    \centering
\includegraphics[width=0.95\textwidth]{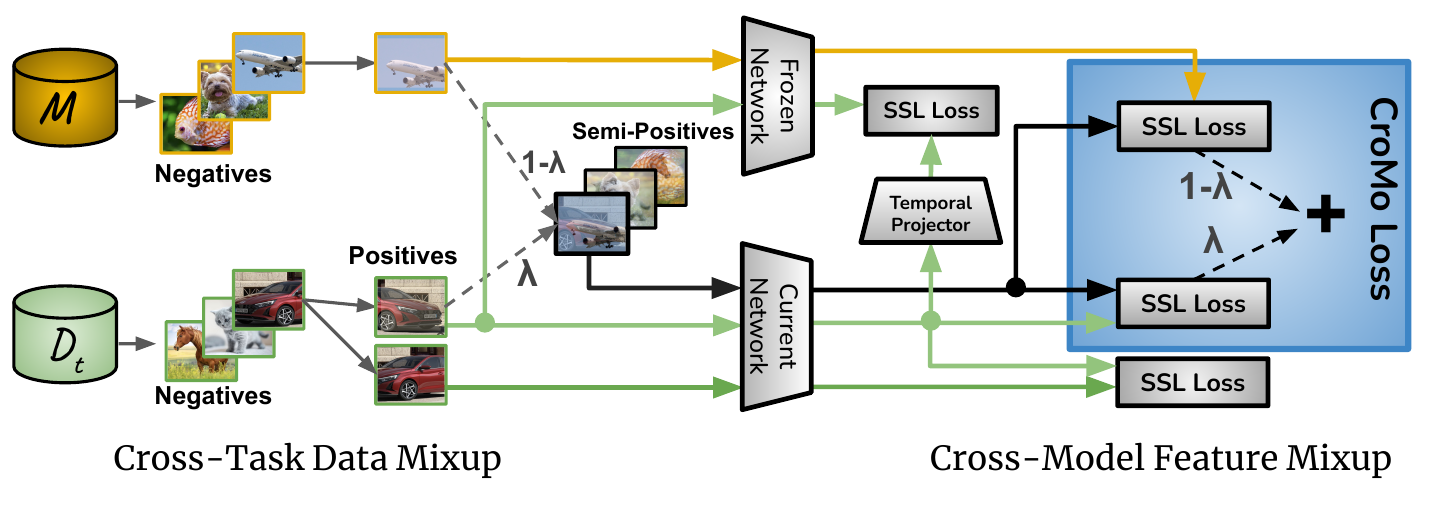}
    \caption{Illustration of our proposed CroMo-Mixup framework. At the input, cross-task mixed samples are generated by a convex interpolation of the current and old task samples from the memory buffer. At the output, the model learns similarities between the embeddings of the cross-task mixed sample and the original samples that were mixed to create it. The embeddings of memory buffer samples come from the frozen network saved from the old task (t-1), whereas mixed samples and current task sample embeddings are attained from the network of the current task (t). In addition, model learns current task via task-specific SSL loss and distills old knowledge on the current task samples via a temporal projector-based distillation loss.}
    \label{fig:systemmodel}

\end{figure}

Continual self-supervised learning (CSSL) refers to the machine learning setting where the model learns tasks sequentially but without data labels. Under continuous shifts in data distributions, deep learning models suffer from catastrophic forgetting; loss of prior tasks knowledge while learning the new tasks. In general, CSSL has two well-explored setups, task-incremental learning (TIL) and class-incremental learning (CIL). In both TIL and CIL, each task has a distinct set of classes. However, for TIL, task id is available at the inference time, whereas, in CIL, evaluation is performed without knowing task-ids. Therefore, CIL is known to be a more challenging setup of continual learning \cite{kim2022theoretical}. 

CSSL has gained research community attention recently, and various knowledge distillation-based methods \cite{cassle, sycon} and exemplar-based algorithms \cite{lump, hu2021well} have been proposed to mitigate catastrophic forgetting. However, a recent study \cite{kim2022theoretical} on continual supervised learning has shown that catastrophic forgetting makes supervised CIL prone to another challenge, named task confusion. In task confusion, as the model learns the new tasks, it may forget the prior tasks knowledge and therefore may fail to establish discriminative decision boundaries between the classes of different tasks. Though they study various supervised CIL works to analyze task-confusion aspect of continual learning (CL), CSSL has not been studied from the perspective of task-confusion before. Motivated by this research gap, we study CSSL from the task-confusion aspect in CIL setups.

First, we hypothesize that the task confusion problem in contrastive SSL methods arises primarily from the inability to train classes belonging to different tasks concurrently. In supervised continual learning, cluster overlap can be a result of forgetting whereas SSL may suffer from this problem even when forgetting effects are eliminated. To demonstrate this, we conduct experiments, presented in Section \ref{motivation}, to study only the task-confusion problem in SSL. Our study shows that contrastive SSL baselines observe a significant drop ($4\%$ or more) in both linear accuracy and task-id prediction when classes are separated across tasks, even if tasks can be revisited frequently. This accuracy drop is compared to the offline setting when classes are randomly sampled from the whole training dataset. The performance drop especially in task-id prediction highlights the model confusion in predicting the task-ids correctly. However, within-task performance remains equally good across both experiment settings. Interestingly, we did not observe such accuracy drop in linear accuracy and task-id prediction for the similar experiment settings of supervised learning, which hints that without forgetting, supervised learning may not experience task confusion.

Given the above-mentioned observations, a straightforward solution can be storing some samples and using them as a replay. However, the challenge is that those limited old task samples might not be enough to create sufficient contrast between classes of old and new tasks as we observed for the ER baseline \cite{er} in Table \ref{task_hetero_tab}, and some other baselines in Table \ref{task_comparison}. Therefore, to integrate memory buffer samples effectively for contrastive learning, we propose Cross-Model feature Mixup (CroMo-Mixup) framework that exploits a small memory buffer and last task's model. As shown in Fig. \ref{fig:systemmodel}, our proposed formulation consists of two components, Cross-Task data Mixup and Cross-Model feature Mixup. Cross-Task data Mixup generates cross-task class mixed samples via mixup data augmentation \cite{zhang2017mixup} to enhance negative sample diversity. Cross-Model feature Mixup formulation learns the similarities between the embeddings of the cross-task mixed samples and the original samples that were mixed to create it. Instead of following the traditional SSL approach of contrasting positives and negatives only, it learns similarities between the original samples and their stochastic mixtures with another negative which can be more challenging. Note that in the proposed formulation, we obtain the embeddings from cross-models, that is, the old task data embedding from the old model and new task data embedding from the new model. This formulation essentially enhances the remembrance of old knowledge via cross-model knowledge retrieval, and learns better class boundaries by learning on diverse and challenging cross-task mixed samples.

Our key contributions in this work are as follows,
\begin{itemize}
    \item[$\diamond$] First, we show the inherent challenges of self-supervised learning that could impact CSSL. With extensive experiments on four SSL baselines, we show the susceptibility of CSSL to the task confusion problem even under relatively simpler setups where forgetting effects are mitigated.
    \item[$\diamond$] We propose a novel cross-model feature mixup framework for CSSL. It creates stochastic mixtures of cross-task data samples that enhance the negative sample diversity. To learn better cross-task class contrast on these samples, we exploit cross-model feature mixup that learns similarities between the cross-model embeddings of the cross-task mixed and original samples. 
    \item[$\diamond$] We implement CroMo-Mixup framework with four SSL baselines, CorInfomax \cite{infomax}, Barlow-Twins \cite{barlowtwins}, SimCLCR \cite{simclr}, and BYOL \cite{byol} to show its compatibility with SSL. For all these baselines, CroMo-Mixup consistently outperforms the state-of-the-art CSSL work, CaSSLe, on three datasets, CIFAR10, CIFAR100, and tinyImageNet. For the best-performing SSL baseline, CroMo-Mixup outperforms  CaSSLe by 4.2\%, 5.8\%, 1.5\%, 5.2\% accuracy improvement on CIFAR100-Split5, CIFAR100-Split10, CIFAR10-Split2 and tinyImageNet-Split10, respectively. 
\end{itemize}
\section{Preliminaries} 

\subsection{Self-Supervised Learning}
Self-supervised learning aims to learn data representations without the need for explicit external labels. Given a dataset \( \mathcal{D} = \{x_1, x_2, ..., x_N\} \), where \( x_i \) represents the \(i\)-th data sample in the dataset, and \( N \) is the total number of samples, the objective of SSL is to learn an embedding function \( f: \mathcal{X} \rightarrow \mathcal{H} \). The embedding function \(f\) maps an input space \( \mathcal{X} \) to a feature space \( \mathcal{H} \) such that 
samples coming from the same class in the input space $\mathcal{X}$ are linearly separable in the embedding space $\mathcal{H}$ from other classes. The most common state-of-the-art SSL methods adopt contrastive learning for this purpose and have shown comparable results to supervised learning \cite{simclr}. These methods use an additional projector network $g$ which is mostly an MLP \cite{barlowtwins, simclr, infomax}. Initially, two different views of input $x_i$ are obtained by applying multiple augmentations $\mathcal{A}$ such as cropping, rotation, color distortion, and noise injection \cite{barlowtwins, simclr, infomax}. The augmented views are regarded as positive pairs for each other. The first view $x_i^1 = \mathcal{A}_1(x_i)$ is fed to the encoder and the projector to yield its representations $z_i^1 = g(f(x_i^1))$; and the second view $x_i^2 = \mathcal{A}_2(x_i)$ is forwarded to the copy of $f$ and $g$ or the target network (e.g same architecture with $f$ and $g$ but parametrized with exponential moving average of parameters of $f$ and $g$) to yield $z_i^2$. Finally, an SSL loss $\mathcal{L}_{SSL}$ is applied between the final features of two views:
\begin{equation}   
\argmin_{\theta} \mathbb{E}[\mathcal{L}_{SSL} (\mathbf{z}^1, \mathbf{z}^2)],
\end{equation}
where $\mathbf{z}^k = [z_1^k, z_2^k, ..., z_N^k, ]$ and $\theta$ represents the parameters of $f$ and $g$ functions together. Some popular SSL loss functions are InfoNCE \cite{simclr, moco}, MSE \cite{byol}, Cross-Correlation \cite{barlowtwins}, Infomax \cite{infomax}. The key objective of these algorithms is to learn distortion-invariant visual representations, i.e., output similar embeddings for the positive pairs, and dissimilar embeddings for the negative samples. We provide further details of our baseline SSL methods in the Appendix \ref{ssl_loss}.

\subsection{Problem Definition and Evaluation Setup}
\subsubsection{Continual Self-Supervised Learning}
We consider Continual Self-Supervised Learning (CSSL) problem, where the main aim is to make neural network continually learn from new data over time without forgetting previously acquired knowledge. Formally, let us consider a sequence of tasks \(\mathcal{T}_1, \mathcal{T}_2, ..., \mathcal{T}_T\) that an SSL model encounters over time, where each task \(\mathcal{T}_t\) is associated with a distinct data \(\mathcal{D}_t = \{(x_{t,i}, y_{t,i})\}_{i=1}^{N_t}\) having \(N_t\) samples and only the corresponding data $\mathcal{D}_t$ is available during task $\mathcal{T}_t$. The goal of CSSL is to optimize the model's performance across all tasks:
\begin{equation}
\argmin_{\theta} \sum_{t=1}^{T} \mathbb{E}[\mathcal{L}_{SSL} (\mathbf{z}_t^1, \mathbf{z}_t^2)],
\end{equation}
where $\mathbf{z}_t^k = [z_{t, 1}^k, z_{t, 2}^k, ..., z_{t, N_t}^k ]$. According to the data distribution across tasks, CSSL can be broadly classified under two setups, Task Incremental Learning (TIL) and and Class Incremental Learning (CIL). In both TIL and CIL, each task has a distinct set of classes. Formally, let $Y_t$ be the set of classes in task $t$, then it is satisfied that $(Y_t \cap Y_{t'}) = \emptyset$ for all $t \neq t'$.  In both cases, new classes occur over time while the data of the previous classes becomes unavailable. However, for TIL, task id is available at the inference time, whereas, in CIL, evaluation is performed without knowing task-ids. We focus on CIL setup as it is often regarded a more challenging setup \cite{kim2022theoretical} due to the unavailability of task-id at the inference time, where the model is expected to differentiate between classes belonging to different tasks. We refer to Class Incremental Self-supervised Learning as CSSL in the rest of this paper.

\subsubsection{Evaluation of Class Incremental Self-Supervised Learning}
Following the setup used in previous CSSL works \cite{cassle}, the performance of an SSL model is measured with linear classification at the end of all tasks while the parameters of encoder network $f$ is frozen. The linear classifier is trained using the set of encoded vectors $\{h_i = f(\mathcal{A}_{lin}(x_i))\}_{i=1}^N$ as inputs, where $\mathcal{A}_{lin}$ is the test data augmentations used in this process (Typically, the training data augmentations $\mathcal{A}_1$ and $\mathcal{A}_2$ are chosen to be harsher than the test data augmentations $\mathcal{A}_{lin}$). After the linear classifier training, the accuracy of the classification on the test dataset is considered as SSL performance.  
To analyze the behaviour of CSSL algorithms better, besides reporting the linear accuracy, we follow \cite{kim2022theoretical} to define and evaluate two sub-problems of CIL in a probabilistic framework. We provide the definition of the two sub-problems, Within-Task Prediction (WP) and Task-ID Prediction (TP), as well as linear accuracy below. 

Let $Y_{t,j}$ be the $j^{th}$  class of $t^{th}$  task, $Y_t$ be the set of all classes at $t^{th}$ task and $X_{t,j}$ be the set of all images belong to $Y_{t,j}$. Linear layer $\phi$ try to map $f(x \in X_{t,j})$ to the $Y_{t,j}$. Following this notation, the metrics are defined below:
\begin{itemize}
    \item \textbf{Linear Accuracy (LA)} is the probability of an image that is correctly classified into its class, i.e, both the task-ID and class within-the-task are correctly classified. Mathematically, LA $= P(\phi(x \in X_{t,j}) = Y_{t,j})$.
    \item \textbf{Task-ID Prediction (TP)} is predicting the task ID. The probability of the linear layer correctly maps an image into one of the classes that belong to the same task of that image. Mathematically,  TP $= P(\phi(x \in X_{t,j}) \in Y_{t})$  
    \item \textbf{Within-Task Prediction (WP)} is predicting the class of an image given the task-id. It is the probability of doing correct classification given that task-id is correctly predicted. Mathematically, WP $= P(\phi(x \in X_{t,j}) = Y_{t,j} | \phi(x \in X_{t,j}) \in Y_{t})$.
\end{itemize}

As it is obviously shown in \cite{kim2022theoretical}, LA = WP $\times$ TP.

\section{Challenges of Class Incremental Self-Supervised Learning}
\label{motivation}
In this section, we explain the challenges of CSSL and their significance in CSSL problem formulation.
\begin{figure}[!b]
    \centering
    \includegraphics[width=0.90\textwidth]{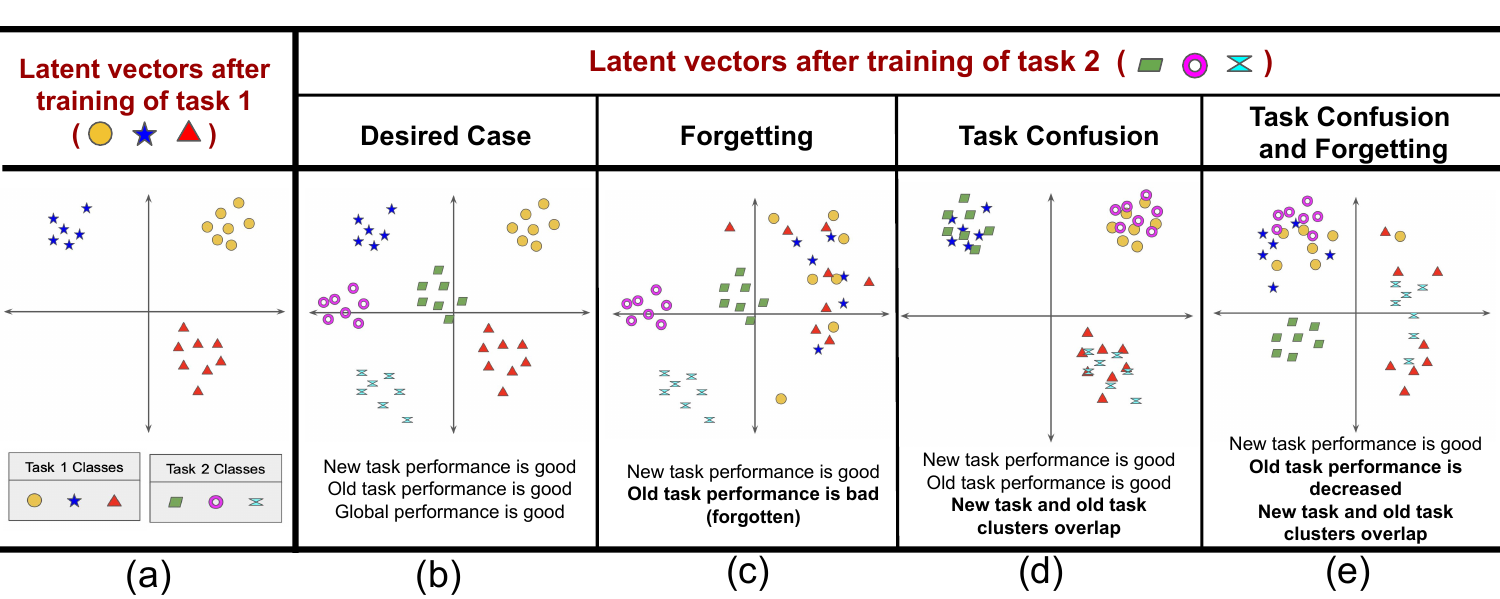}
    \caption{Demonstration of Catastrophic Forgetting and Task Confusion challenges in a two-task based Continual Learning setup where each task contains three classes. Figure (a) illustrates the linear separability of latent vectors of task 1 classes at the end of task 1 training. Figures (b)-(e) represent the four cases after training on task 2. Case (b) shows the desired case where all classes of both tasks are linearly separable. Figure (c) illustrates the forgetting effect where task 2 classes are linearly separable but task 1 classes are not. Figure (d) shows the task confusion problem, where the model fails to draw distinctive decision boundaries between different task classes and may have overlapping clusters. Figure (e) shows the effects of task confusion and forgetting together, which is the problem in CSSL settings we want to solve. }
    \label{fig:task_confusion_illustration}
\end{figure}
\subsection{Catastrophic Forgetting} Catastrophic forgetting is the most addressed issue in continual self-supervised learning literature. It represents the significant loss of performance on previous tasks upon learning new ones. Due to forgetting of the previous tasks, the model's within-task prediction performance on the previous tasks decreases substantially, whereas it remains at a desirable level on the current task. Forgetting also degrades the task-id prediction performance as shown in Figure \ref{fig:task_confusion_illustration}. The earlier CSSL works exploited memory replay to address catastrophic forgetting  \cite{hu2021well}. The state-of-the-art CSSL works have proposed self-supervised learning loss adaptation for knowledge distillation \cite{cassle, sycon} and fine-tuning \cite{tang2024kaizen} to mitigate this problem. However, the current literature on CSSL does not pay due attention to task confusion challenge that can hinder learning distinctive representations in the absence of labels as explained below.
\subsection{Task Confusion} 
Task confusion represents the model failure to establish distinctive decision boundaries between different classes belonging to different tasks \cite{kim2022theoretical, huang2023resolving}. Task confusion is crucial in CIL because the task id is not present at the inference time and its absence could result in the learner's failure to accurately predict the task id, leading to mis-classification of test images. 
In supervised continual learning, task confusion arises as a result of forgetting which leads to the overlap of inter-task class embeddings clusters as shown in Figure \ref{fig:task_confusion_illustration} (subfigure (e)). However, in this work, we show that contrastive learning-based SSL methods can be susceptible to inter-task class separation problem even when forgetting effects are eliminated such as shown in Figure \ref{fig:task_confusion_illustration} (subfigure (d)). The model remembers old samples and WP is good, however, the model struggles to identify task-id correctly due to clusters overlap. This is because class-incremental setup naturally leads to a lesser diversity of negative samples as old task data cannot be visited with new task data to draw a contrast in the absence of labels. To illustrate it further, we describe our hypothesis and experiment results below.

Our hypothesis is as follows,
\begin{adjustwidth}{0.9cm}{0.9cm}
  {\textit{The task confusion problem in Contrastive SSL methods arises primarily from the inability to train the model with different classes belonging to different tasks concurrently.}}
\end{adjustwidth}
\begin{figure}[b!]
    \centering
    \includegraphics[width=0.9\textwidth]{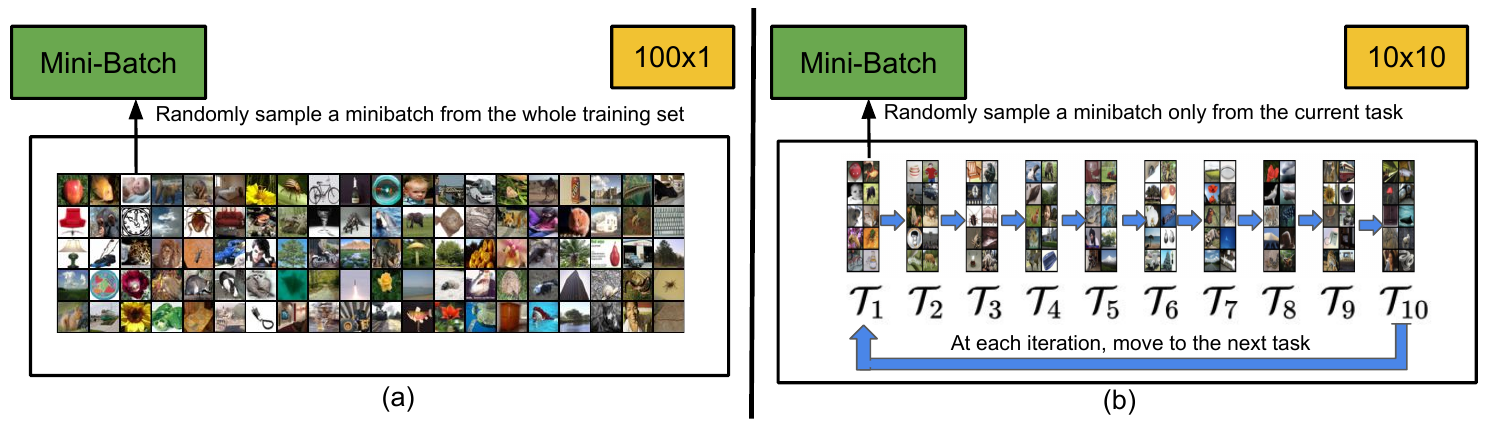}
    \caption{Depiction of 100x1 and 10x10 CIL-minibatch task confusion experiment setup on the CIFAR100 dataset. Figure (a) represents the 100x1 case where a regular uniform sampling is performed from all the samples containing all 100 classes. Figure (b) shows the 10x10 setting where there are 10 tasks and each task contains only 10 classes. Classes are mutually exclusive across tasks. For SSL training, a mini-batch is sampled only from a single task at a time. After each iteration, mini-batch sampler moves to the next task so that tasks can be revisited throughout the training. }
    \label{fig:bw}
\end{figure}
To study the hypothesis, we explore both self-supervised and supervised learning in a fairly simple but representative class-incremental setup. Our experiment setup is as follows: we follow a traditional CIL setup with a sequence of tasks \(\mathcal{T}_1, \mathcal{T}_2, ..., \mathcal{T}_T\) that are mutually exclusive in classes. We consider the CIFAR100 dataset and split 100 classes across 10 tasks by assigning 10 classes per task. Further, we assume that tasks change after each iteration (one gradient descent step), i.e., mini-batches are sampled from different tasks at each iteration as shown in Figure \ref{fig:bw}. To study the task confusion problem explicitly and remove the forgetting effect, we assume that tasks can be revisited, i.e., task 2 follows task 1, task 3 follows task 2, and so on. The repeatability of tasks ensures that the SSL learner does not forget the previous knowledge while mutual exclusivity of classes is also maintained across mini-batches naturally leading to the representative CIL setting of lesser diversity across mini-batches. For simplicity, we refer to this experimental setup, 10x10 class-incremental learning across mini-batches, 10x10 CIL-minibatch because the data is divided into 10 tasks where each task contains 10 classes. To show how the performance of the methods changes in CIL-minibatch setting, we also do a training on the regular setting where we sample uniformly random from the whole training data. We call this regular training setting as 100x1 CIL-minibatches because there is 1 task containing all 100 classes. Lastly, we focus on the training accuracy of the methods because we only care about the methods' capability of creating linearly separable features on the data they are trained on. We present our key results in Fig. \ref{fig:bw_results}. Our main observations from the results of these experiments are:
\begin{figure}[b!]
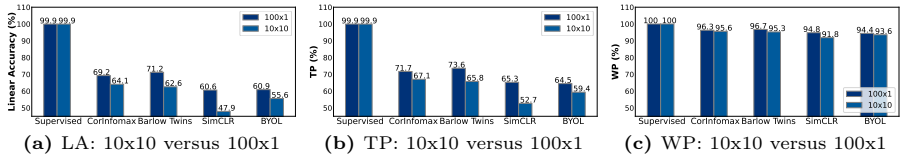

\centering
\begin{subfigure}{.32\textwidth}
    \centering
    \includegraphics[width=.98\linewidth]{figures/CIL_train_lin_update.pdf}  
    \caption{LA: 10x10 versus 100x1}
    \label{SUBFIGURE LABEL 1}
\end{subfigure}
\begin{subfigure}{.32\textwidth}
    \centering
    \includegraphics[width=.98\linewidth]{figures/CIL_train_tp__update.pdf}  
    \caption{TP: 10x10 versus 100x1}
    \label{SUBFIGURE LABEL 2}
\end{subfigure}
\begin{subfigure}{.32\textwidth}
    \centering
    \includegraphics[width=.98\linewidth]{figures/CIL_train_wp_update.pdf}  
    \caption{WP: 10x10 versus 100x1}
    \label{}
\end{subfigure}
\caption{Training LA, WP, and TP performance of contrastive SSL methods, CorInfomax \cite{infomax}, Barlow-Twins \cite{barlowtwins}, SimCLR \cite{simclr}, and BYOL \cite{byol}, and supervised learning on the CIFAR100 Dataset for 100x1 and 10x10 CIL-minibatch settings. Figure (a) demonstrates that the 10x10 setting leads to a significant accuracy drop across all SSL baselines as compared to the 100x1 setting. Figure (b) presents that the lower linear accuracy is reflected in lower task-id prediction performance, demonstrating the task-confusion problem. Figure (c) shows that the WP performance remains relatively good. }
\label{fig:bw_results}
\end{figure}
\begin{itemize}
    \item \textbf{1}: The linear accuracy of all four representative self-supervised learning models drops significantly in 10x10 CIL mini-batch experiments as compared to the 100x1 Joint-SSL case. The accuracy problem stems from the fact that certain classes are not trained in the same mini-batch together. The lower linear accuracy is reflected majorly in lower TP whereas WP remains good overall. This hints that when trained in a lesser negative diversity setup such as CIL, self-supervised learning suffers from the task confusion problem as reflected in lower Task-ID prediction performance. 
    \item \textbf{2}: In contrast to self-supervised learning, supervised learning exhibits no change in accuracy and maintains linear separability of classes in the embedding space for both 100x1  joint-SL and 10x10 test cases. This shows that this challenge is unique to contrastive learning-based SSL methods and may not affect supervised learning due to the presence of explicit class labels.
\end{itemize}
Overall, our experimental results in this section confirm our hypothesis, underlining the importance of addressing task confusion problem in CSSL. A data incremental setting-based ablation study that further strengthens our hypothesis can be found in Appendix Section \ref{task_cobfusion_hps}.

\section{Proposed Method}
In CSSL, the main objective is to learn visual representations that remain informative about the old task data distributions while learning the new task such that linear separability of all the classes from all the data distributions is maximized at the end of the CL Phase. Existing works have proposed supervised learning solution adaptations to CSSL to address catastrophic forgetting such as Distillation \cite{cassle, sycon}, and memory-replay \cite{lump}. However, the CSSL representation continuity has not been studied from task confusion perspective before. As we have shown in Section \ref{motivation}, task confusion is a major challenge for SSL. In literature, it is well-known that SSL often requires large unsupervised datasets, and larger mini-batches to ensure sufficient negative sample diversity from all the classes to learn the linear separability of different classes in the embedding space \cite{simclr, byol, wang2020understanding}. However, under the CSSL problem setup, only a small amount of data can be saved (as an exemplar), and this might not be sufficient to produce the desired negative sample diversity even when used as a replay (as shown in the results Section, e.g., \ref{results}, ER \cite{er}, DER \cite{der}, EWC \cite{EWC}, LUMP \cite{lump}). Given these challenges, we propose an exemplar-based approach that focuses on enhancing the negative sample diversity under the limited memory buffer constraints.

Our proposed framework consists of two components: 1) Cross-Task Data Mixup and 2) Cross-Model Feature Mixup which are described in detail below.

\subsubsection{Cross-Task Data Mixup} 
We generate cross-task mixed samples by exploiting the well-known mixup data augmentation \cite{zhang2017mixup} for self-supervised learning. Specifically, for $x_{{t}_{i}}$ sampled from current task data distribution $\mathcal{D}_{t}$, we randomly sample $x_{\mathcal{M}_{j}}$ from the memory buffer $\mathcal{M}$ that contains samples from the previous task data distributions, and generate inter-task class mixed data sample $x_{{mix}_{ij}}$, a convex interpolation of $x_{{t}_{i}}$ and $x_{\mathcal{M}_{j}}$ as shown below,
\begin{equation}
\label{mixup}
    x_{{mix}_{ij}} = \lambda x_{{t}_{i}} + (1-\lambda) x_{\mathcal{M}_{j}}
\end{equation}
where $\lambda \in \text{Beta}(\alpha, \alpha)$, and $\alpha \in (0, \infty)$. Note that we use mixup for both views of data samples, $(x^{1}_{{t}_{i}}, x^{2}_{{t}_{i}}$) and $(x^{1}_{\mathcal{M}_{j}},x^{2}_{\mathcal{M}_{j}})$ which subsequently results in two inter-task mixed data samples $x^{1}_{{mix}_{ij}}$ and $x^{2}_{{mix}_{ij}}$. 
\subsubsection{Cross-Model Feature Mixup (CroMo-Mixup)} For learning on the cross-task mixed data samples, our proposed formulation is as follows,
\begin{equation}
\label{cromo-mixup}
    \mathcal{L}_{CroMo}(z_{{mix}_{ij}}, z_{{t}_{i}}, \bar{z}_{\mathcal{M}_{j}}) = \lambda \cdot \mathcal{L}_{SSL}(z_{{mix}_{ij}}, z_{{t}_{i}}) + (1-\lambda) \cdot \mathcal{L}_{SSL}(z_{{mix}_{ij}}, \bar{z}_{\mathcal{M}_{j}})
\end{equation}
where $\mathcal{L}_{SSL}$ is the SSL loss for the considered SSL baseline. $z_{{mix}_{ij}}$, $z_{{t}_{i}}$ are the feature embeddings obtained from the current model for the $x_{{mix}_{ij}}$ and $x_{{t}_{i}}$ data samples, and $\bar{z}_{\mathcal{M}_{j}}$ feature embedding is obtained from the frozen old task model for the $x_{\mathcal{M}_{j}}$ data-point. 

The proposed learning objective has three key features. For a given $x_{{t}_{i}}$, $x_{\mathcal{M}_{j}}$, and $x_{{mix}_{ij}}$ data samples, it treats the embeddings of the rest of the cross-task mixed samples $z_{mix}$, the current task embeddings $z_{t}$, and the old task embeddings $\bar{z}_{\mathcal{M}}$ in the mini-batch as negatives for each other which enriches the negative sample diversity of overall learning, as compared to the traditional SSL learning where we only have the old and new task's data embeddings as negatives for each other. Second, it encourages the learner to learn the similarities between the cross-task mixed sample $z_{{mix}_{ij}}$ and the corresponding current task sample $z_{{t}_{i}}$ as well as the old task sample $\bar{z}_{\mathcal{M}_{j}}$. This soft distance learning helps in improving the task-id prediction performance because learner learns to identify the similarity of an image from a more challenging image, augmented as well as mixed with a cross-task negative sample, than only an augmented version of itself. Further, the proposed formulation exploits the old task data embeddings from the old model $\bar{z}_{\mathcal{M}}$, which promotes retrieval and preservation of the old knowledge while learning new knowledge. 

In addition to the learning objective \ref{cromo-mixup}, a general task-specific loss is also employed $\mathcal{L}_{SSL}$ to learn the current task on the current task data distribution $\mathcal{D}_{t}$. We also exploit distillation on the current task data distribution $\mathcal{D}_{t}$ benefiting from the old task model. Hence, the total objective becomes,
\begin{equation}
\begin{split}
\mathcal{L}_{total}&= \mathcal{L}_{SSL}({z}^{1}_{{t}_{i}}, z^{2}_{{t}_{i}})  + \zeta(\mathcal{L}_{SSL}(\bar{z}^{1}_{{t}_{i}}, h({z}^{1}_{{t}_{i}})) + \mathcal{L}_{SSL}(\bar{z}^{2}_{{t}_{i}}, h({z}^{2}_{{t}_{i}}))) + \\& \mathcal{L}_{CroMo}(z^{1}_{{mix}_{ij}}, z^{1}_{{t}_{i}}, \bar{z}^{1}_{\mathcal{M}_{j}}) +\mathcal{L}_{CroMo}(z^{2}_{{mix}_{ij}}, z^{2}_{{t}_{i}}, \bar{z}^{2}_{\mathcal{M}_{j}}) 
    \end{split}
\label{totalloss}
\end{equation}
where $\zeta$ is a hyper-parameter for the distillation objective. Further, $\mathcal{L}_{SSL}({z}^{1}_{{t}_{i}}, z^{2}_{{t}_{i}})$ denotes a task specifc loss. $h(.)$ represents an MLP Predictor that is employed on $z^{2}_{{t}_{i}}$ and $z^{1}_{{t}_{i}}$ to perform distillation as proposed in \cite{cassle}. The embeddings $\bar{z}^{1}_{{t}_{i}}$ and $\bar{z}^{2}_{{t}_{i}}$ are obtained from the old model for the inputs $x^{1}_{{t}_{i}}$ and $x^{2}_{{t}_{i}}$. 

It is worth mentioning that mixup \cite{lee2020mix, kalantidis2020hard,kim2020mixco} has already been explored in contrastive SSL works as a data augmentation scheme, however, here we exploit it to formulate a CSSL problem and perform learning in a cross-task and cross-model continual learning setting. Further, one baseline, LUMP \cite{lump}, has used the idea of mixup for CSSL, however, our proposed objective formulation is different from theirs. Primarily, the formulation proposed in LUMP exploits cross-task mixed samples such that it learns general-purpose features across tasks to address forgetting. Whereas our objective function learns to identify task-specific features at a granular level to address both task confusion and forgetting problems in CSSL. LUMP minimizes the SSL loss over the mixed samples and their augmented versions, whereas CroMo-Mixup learns to find the feature similarity between the original and cross-task mixed samples in the same proportion in which they were mixed. This implicit feature learning ensures that the model can identify different task samples at a granular level. Further, CroMo-Mixup outperforms this baseline in CSSL settings as shown in Table \ref{task_comparison}.

\section{Related Works}
\label{related_works_lump}
\noindent \textbf{Continual Self-Supervised Learning.} 
The research community has recently shown keen interest in CSSL problem \cite{lump, sycon, hu2021well, cassle} due to its applicability in real-world scenarios. In this line of research, \cite{lump} is among the first works that demonstrated the representation continuity of SSL in task-incremental learning settings. It proposed a cross-task data mixup approach and showed that its method outperforms various supervised learning baselines in TIL settings. Another work \cite{cheng2023contrastive}, has explored CSSL for TIL settings where task confusion is not a concern. Other works \cite{lin2022continual, hu2021well} have investigated the significance of simple memory replay to address catastrophic forgetting in CSSL. However, CaSSLe \cite{cassle} made significant progress. They proposed self-supervised learning loss function adaptation via a temporal projector to perform distillation. Sy-Con \cite{sycon}, another recent work, proposed a loss formulation that exploits current and old model embeddings of negative samples to enhance distillation regularization performance. Nonetheless, CaSSLe remains state-of-the-art on most self-supervised learning baselines. Due to space constraints, we present the literature review of SSL and Continual Learning topics in the Appendix \ref{relatedworks}.

\section{Experiments}
\label{results}
\subsection{Experiment Settings}
\subsubsection{Datasets}
We perform experiments on three datasets: CIFAR10 \cite{krizhevsky2009learning}, a 10-class dataset with 60,000 32x32 color images; CIFAR100 \cite{krizhevsky2009learning}, a 100-class dataset with 60,000 32x32 color images; and TinyImageNet \cite{le2015tiny}, a 200-class dataset with 100000 64x64 color images. For CIFAR10, we explore a 2 task setting where 5 classes are present per task. Following \cite{cassle}, we experiment with a 5-task class-incremental setting for CIFAR100. Further, we also include a more challenging case with 10 tasks for CIFAR100 dataset. For tinyImageNet, we exploit a 10-task setting where 20 classes are present per task. We provide further details for each dataset setup in Appendix Section \ref{HPs}.
\label{HPs}

\subsubsection{Implementation Details}
We use ResNet-18 \cite{he2016deep} as an encoder network for CIFAR10 and CIFAR100 experiments, while we employ ResNet-50 \cite{he2016deep} for TinyImageNet. We include BYOL \cite{byol}, SimCLR \cite{simclr}, CorInfoMax \cite{infomax}, and Barlow-Twins \cite{barlowtwins} as representative SSL baselines in our work.  We follow these works to set up the hyper-parameters such as optimizer, learning rate, and schedulers. For Continual Learning experiments, we use 500 epochs/task for CIFAR10-Split2, 750 epochs/task for CIFAR100-Split5, 600 epochs for the first task, and 350 epochs for the rest of the task for CIFAR100-Split10. Likewise, for tinyImageNet, we use 500 epochs for the first task and 350 epochs for the rest of the tasks. Further details of hyper-parameters tuning can be found in Appendix \ref{HPs}.
\subsubsection{Evaluation Metrics}
Following the CSSL baseline \cite{cassle}, we evaluate the model at the end of CL training. To evaluate the model, we freeze the encoder and train a linear classifier layer on the training dataset of each specific dataset. We report the average linear accuracy on the test set of each specific dataset, which is calculated as $=\frac{\text{Total \# of Correct Classification of both Class and Task-ID Prediction Tasks}}{\text{Total \# of Test samples}}$. To analyze the model performance against task confusion, we also report TP, which is calculated as ($\frac{\text{Total \# of Correct Classification of Task-ID Prediction Task}}{\text{Total \# of test samples}}$). For WP, we report, ($\frac{\text{Total \# of Correct Classification of both Class and Task-ID Prediction Tasks}}{\text{Total \# of Correct Classification of Task-ID Prediction Task}}$). 

\subsection{Results} 
\begin{table}[b!]
\centering
\caption{Experimental results of CSSL baselines on CIFAR100-Split5}
\fontsize{6}{8}\selectfont
\begin{tabular}{|c|c|c|c| }
\hline
\multicolumn{1}{|c|}{} & \multicolumn{1}{c|}{\textbf{Barlow-Twins}} & \multicolumn{1}{c|}{\textbf{SimCLR}} & \multicolumn{1}{c|}{\textbf{BYOL}} \\

\hline
Method & \textbf{Avg. Linear Acc(\%)} & \textbf{Avg. Linear Acc(\%)} & \textbf{Avg. Linear Acc(\%)}\\

\hline

Offline     & 70.0   & 65.1 & 66.7 \\
Fine-Tune     & 54.4   & 42.7 & 55.2 \\
$\text{EWC}^{\dagger}$    & 56.7   & 53.6 & 56.4 \\
ER     & 57.2   & 47.7 & 56.1 \\
$\text{DER}^{\dagger}$     & 55.3   & 50.7 & 54.8 \\
$\text{LUMP}^{\dagger}$     & 57.8   & 52.3 & 56.4 \\
Sy-Con  & 60.4  & 58.9 & 57.3 \\
CaSSLe   & 60.6   & 57.6 & 56.9 \\
CaSSLe+   & 61.3   & 59.5 & 57.4 \\
CroMo-Mixup  &  65.5(\textbf{+4.2}) & 62.7(\textbf{+3.2})& 60.6(\textbf{+3.2}) \\

\hline
\end{tabular}
\label{task_comparison}
\end{table}

\subsubsection{Average Linear Accuracy}
First, we evaluate CroMo-Mixup with Barlow-Twins, SimCLR, and BYOL on the CIFAR100-Split5. We compare the average linear accuracy performance of CroMo-Mixup with CSSL baselines CaSSLe \cite{cassle}, Sy-Con \cite{sycon}, and LUMP \cite{lump}, and some replay-based methods from supervised continual learning that can be adapted to CSSL such as EWC \cite{EWC}, ER \cite{er}, DER \cite{der} on CIFAR100-Split5 dataset in Table \ref{task_comparison}. We also include CaSSLe+ baseline that exploits both knowledge distillation and memory buffer to make a fair comparison with the state-of-the-art baseline, CaSSLe. Overall, CroMo-Mixup outperforms all these baselines by achieving higher average linear accuracy across all three SSL baselines. Further, though we observe 1-2\% accuracy gain in CaSSLe+ as compared to CaSSLe, our proposed method outperforms this SOTA baseline, CaSSLe+, by $4.2\%$, $3.2\%$, and $3.2\%$ accuracy improvement on Barlow-Twins, SimCLR, and BYOL, respectively. In Table \ref{task_comparison}, we reproduce all baseline results except {$\dagger$} marked which we take from CaSSLe paper \cite{cassle}.

\begin{table}[h]

\caption{Comparison of CroMo-Mixup with state-of-the-art CSSL method, CaSSLe on CIFAR10-Split2, CIFAR100-Split5, CIFAR100-Split10 and tinyImageNet-Split20 using average linear accuracy (LA), within-task prediction (WP) \& Task-ID prediction (TP)}
\centering
\fontsize{4.5}{8}\selectfont
\begin{tabular}{ c|l |ccc |ccc |ccc |ccc|}
\hline
& & \multicolumn{3}{|c}{\textbf{Barlow-Twins}} & \multicolumn{3}{|c}{\textbf{CorInfoMax}} & \multicolumn{3}{|c}{\textbf{SimCLR}} & \multicolumn{3}{|c|}{\textbf{BYOL}} \\
 \hline

& Method & LA(\%) & WP(\%) & TP(\%) & LA(\%) & WP(\%) & TP(\%)  & LA(\%) & WP(\%) & TP(\%) & LA(\%) & WP(\%) & TP(\%) \\

\hline
\hline
\multirow{7}{*}{\rotatebox{90}{CIFAR10-Split2}}
&Offline     & 91.65 & -     & -     & 92.18 & -& - & 90.35 & -& - & 89.60 & - & -\\
&Fine-tune   & 82.67 & 90.12 & 91.73 & 81.71 & 91.25 & 89.55 & 80.97 & 90.54 & 89.43 & 84.16 & 94.43 & 89.12\\
&ER          & 85.61 & 90.36 & 94.62 & 87.67 & 95.81 & 91.50 & 81.52 & 90.95 & 89.63 & 86.71 & 94.97 & 91.30\\
&CaSSLe      & 87.64 & 91.25 & 95.87 & 87.62 & 96.02 & 91.17 & 86.88 & 95.21 & 91.25 & 87.00 & 95.80 & 90.81\\
&CaSSLe+ & 86.81 & 90.30 & 95.91 & 87.58 & 95.51 & 91.70 & 87.52 &95.54 & 91.61& 87.81  & 96.01  & 91.49\\
&CroMo-Mixup*& 87.56 & 92.11 & 94.69 & 86.00 & 93.54 & 91.96 &84.18 & 92.70& 90.81& 89.27 & 96.28 & 92.72\\
&CroMo-Mixup& 88.22& 91.78 & 95.75 & 88.51 & 95.97 & 92.23 & 88.49 & 95.93 & 92.24& 88.88 & 96.36 & 92.19\\
& & (\textbf{+0.6})&  & (\textbf{-0.1}) & (\textbf{+0.8}) &  & (\textbf{+0.7}) & (\textbf{+1.0}) & -- & (\textbf{+0.6})& (\textbf{+1.5}) & -- & (\textbf{+0.7})\\

\hline
\hline
\multirow{8}{*}{\rotatebox{90}{CIFAR100-Split5}}
&Offline     & 70.03 & -& -    & 70.76 & -& - & 65.11 & -& - &66.73 & - & -\\
&Fine-tune   & 54.40 & 85.19&63.86  &  56.68 & 86.32&65.66 & 42.65 & 78.17&54.56 &55.19 & 86.23 & 64.00\\
&ER          & 57.23 &88.26 &65.57 &59.94 &88.62 &67.64  & 47.77 & 81.87 & 58.35 & 56.05 & 85.60 & 65.48\\
&CaSSLe      & 60.64 &87.29 & 66.35& 60.82&88.85 & 68.45 & 57.54&87.89 & 65.47& 56.86 & 86.05 & 66.08\\
&CaSSLe+ & 61.25& 88.50& 69.03 & 60.26&88.74& 67.92 &59.48 &88.26 & 67.39 & 57.35 & 87.28 & 65.71 \\
&CroMo-Mixup*& 63.94 & 91.40 & 69.88  &62.32 & 88.46 & 70.39& 59.15 & 87.66 & 67.48&59.60 & 88.11 & 67.64\\
&CroMo-Mixup& 65.48 & 90.72 &72.11& 65.06 & 90.62 &71.78 & 62.72 & 89.50&70.08 &60.60 & 88.64 & 68.37\\
&& (\textbf{+4.2}) &  &(\textbf{+3.1})& (\textbf{+4.2}) &  &(\textbf{+3.3}) &  (\textbf{+3.2})&& (\textbf{+2.7}) &(\textbf{+3.3})& -& (\textbf{+2.7})\\
\hline
\hline

\multirow{8}{*}{\rotatebox{90}{CIFAR100-Split10}}
&Offline     & 70.03 &- &-    & 70.76 & -& - & 65.11 & - & - &66.73& - & -\\
&Fine-tune   & 51.12&92.01&55.56 & 50.66&91.58&55.32 &39.02&86.61&45.04 & 49.63 & 92.23 & 53.81\\
&ER          & 52.81 & 92.59 & 56.98 &56.99 & 93.98 & 60.64& 44.83 & 89.98 & 49.88 & 52.32 & 92.70 & 56.40\\
&CaSSLe      & 56.59 & 93.93 & 60.25 &56.35 & 93.95 & 59.98& 53.60 & 93.39 & 57.38 & 52.77 & 93.22 & 56.61\\
&CaSSLe+&56.64 & 93.49&60.68 &57.00 & 93.95&60.67& 55.02 & 93.43&58.89 &53.39 & 92.67 & 57.61\\
&CroMo-Mixup*&60.01 &94.50 & 63.41 &60.30 &94.34 & 63.74& 55.21 &92.84 & 59.84 &56.59 & 93.52 & 60.51\\
&CroMo-Mixup&62.48&95.10&65.70 &61.66&94.91&64.97& 58.84&94.66&62.18 &56.97 & 93.35 & 61.03\\
& &(\textbf{+5.8})&&(\textbf{+5.0}) &(\textbf{+4.7})& &(\textbf{+4.3})& (\textbf{+3.8})&& (\textbf{+3.3}) & (\textbf{+3.6}) & - &  (\textbf{+3.4})\\
\hline
\hline
\multirow{8}{*}{\rotatebox{90}{tinyImageNet-Split10}}
&Offline     & 55.60 & -& - & 55.20 & -& -& 49.74  & -  & - &  47.58& - & -\\
&Fine-tune   & 39.90 & 77.00 & 51.82 & 41.14  & 78.12 & 52.66 & 36.72 & 75.09 & 48.90 & 37.15 & 76.10 & 48.82 \\
&ER        & 40.14 & 77.07 & 52.08 & 41.44 & 78.22 & 52.98 & 37.96 & 70.08 & 50.56 & 37.78 & 76.17 & 49.60\\
&CaSSLe & 43.40 & 79.08 & 54.88  & 41.66 & 77.69 & 53.62 & 40.66 & 77.80 & 52.26 & 38.18 & 77.07 & 49.54\\
&CaSSLe+& 42.64 & 78.90 & 54.04 & 43.86 & 79.46 & 55.20 & 41.74 & 78.58 & 53.12 & 40.24 & 78.72 & 51.12\\
&CroMo-Mixup*& 45.70 & 80.54 & 56.74 & 46.74 & 81.32 & 57.48 & 41.02 & 78.43 & 52.30 & 43.19 & 79.45 & 54.36\\
&CroMo-Mixup& 47.32 & 81.78 & 57.86 & 48.22 & 81.90 & 58.88 & 45.82 & 80.36 & 57.02 & 45.44 & 80.68 & 56.32\\
&& (\textbf{+3.7}) &  & (\textbf{+3.8}) &(\textbf{+4.4}) &  & (\textbf{+3.6}) & (\textbf{+4.1}) &  & (\textbf{+3.9}) & (\textbf{+5.2}) &  & (\textbf{+4.2})\\
\hline
\end{tabular}
\label{task_hetero_tab}
\end{table}
Next, we draw a more detailed comparison between the proposed formulation and state-of-the-art baseline, CaSSLe and its variant CaSSLe+. We perform experiments across four CIL setups, three datasets, and four SSL baselines, represented in Table \ref{task_hetero_tab}. We also include a memory replay based baseline ER to show that a limited memory buffer may not be enough to ensure sufficient negative sample diversity for optimal CSSL performance. For our proposed formulation, we include two setups, $\zeta = 0$ shown as CroMo-Mixup* in Table \ref{task_hetero_tab}, that uses cross-model learning but does not exploit knowledge distillation, and $\zeta = 1$, shown as CroMo-Mixup that exploits both cross-model learning as well as knowledge distillation. On CIFAR10-Split2 dataset, we achieve the highest performance of 89.27\% with CroMo-Mixup on the BYOL baseline that outperforms CaSSLe+ with 1.5\% higher average linear accuracy. Next, for both CIFAR100-Split5 and CIFAR100-Split100 datasets, CroMo-Mixup achieves the highest average linear accuracy of 65.48\% and 62.48\%,  respectively, on Barlow-Twins among all four SSL baselines. It outperforms CaSSLe+ with 4.2\% and 5.8\% higher average linear accuracy performance on CIFAR100-Split5 and Split10, respectively. It is worth noticing that even without distillation, CroMo-Mixup*($\zeta =0$) case, CroMo-Mixup outperforms CaSSLe+, which exploits both knowledge distillation and memory buffer, with 2.7\% and 3.4\% higher linear accuracy performance. On tinyImageNet-Split10, CroMo-Mixup on BYOL achieves a higher accuracy of 45.44\% outperforming CaSSLe+ with a 5.2\% accuracy gain.

\subsubsection{Task-ID Prediction}
To analyze performance against the task-confusion, we also compare the Task-ID prediction performance of CroMo-Mixup with CaSSLe and its variant CaSSLe+ across four CIL setups, three datasets, and four SSL baselines in Table \ref{task_hetero_tab}. We observe that CroMo-Mixup achieves better performance in predicting task-ids as compared to CaSSLe and CaSSLe+ without compromising on the WP performance on nearly all SSL baselines and all considered dataset settings. The higher linear accuracy indicates the model performance against catastrophic forgetting, whereas better task-id prediction performance indicates the potential of CroMo-Mixup to maximize contrast between the classes of different tasks in the absence of class labels and limited buffer size. We also include further experiments such as an ablation study on different design components, out-of-distribution performance comparison, and buffer size versus accuracy analysis in Appendix \ref{further_experiments}.

\section{Conclusion}
In this work, we study continual self-supervised learning (CSSL) from the task-confusion aspect of continual learning. First, we highlight its significance in CSSL problem which remained unexplored before in literature. Next, we propose a CroMo-Mixup formulation that exploits cross-task data mixup and cross-model feature mixup to enhance negative sample diversity and cross-task class contrast under limited memory buffer constraint of continual learning. Our proposed formulation outperforms the state-of-the-art CSSL works by achieving higher performance average linear accuracy and task-id prediction performance.

\section{Limitations}
Our proposed approach is based on a limited memory buffer. Therefore, it may not be applicable in scenarios where the user might want to delete all old samples due to privacy issues. In such cases, there is a need to design effective methods to address the task confusion challenge of the continual self-supervised learning problem. Further, we assume the tasks are clearly separated following the current literature \cite{cassle, sycon, lump}. However, in realistic scenarios, the transitions across tasks are mostly smoother. Therefore, this can be an important future work to explore.

\section*{Acknowledgements}
This work is supported in part by a research gift from USC-Amazon Center on Secure and Trusted Machine Learning \footnote{https://trustedai.usc.edu}, ONR grant N00014-23-1-2191, and ARO grant W911NF-22-1-0165. The work of Jie Ding was supported in part by the Army Research Office under the Early Career Program Award, Grant Number W911NF-23-10315.

%
%
\bibliographystyle{splncs04}
\bibliography{egbib}

\begin{thebibliography}{10}
\providecommand{\url}[1]{\texttt{#1}}
\providecommand{\urlprefix}{URL }
\providecommand{\doi}[1]{https://doi.org/#1}

\bibitem{bakman2024federated}
Bakman, Y.F., Yaldiz, D.N., Ezzeldin, Y.H., Avestimehr, S.: Federated orthogonal training: Mitigating global catastrophic forgetting in continual federated learning. In: The Twelfth International Conference on Learning Representations (2024), \url{https://openreview.net/forum?id=nAs4LdaP9Y}

\bibitem{bardes2022vicreg}
Bardes, A., Ponce, J., LeCun, Y.: {VICR}eg: Variance-invariance-covariance regularization for self-supervised learning. In: International Conference on Learning Representations (2022), \url{https://openreview.net/forum?id=xm6YD62D1Ub}

\bibitem{der}
Buzzega, P., Boschini, M., Porrello, A., Abati, D., Calderara, S.: Dark experience for general continual learning: a strong, simple baseline. Advances in neural information processing systems  \textbf{33},  15920--15930 (2020)

\bibitem{swav}
Caron, M., Misra, I., Mairal, J., Goyal, P., Bojanowski, P., Joulin, A.: Unsupervised learning of visual features by contrasting cluster assignments. Advances in neural information processing systems  \textbf{33},  9912--9924 (2020)

\bibitem{dino}
Caron, M., Touvron, H., Misra, I., J{\'e}gou, H., Mairal, J., Bojanowski, P., Joulin, A.: Emerging properties in self-supervised vision transformers. In: Proceedings of the IEEE/CVF international conference on computer vision. pp. 9650--9660 (2021)

\bibitem{sycon}
Cha, S., Cho, K., Moon, T.: Augmenting negative representation for continual self-supervised learning. https://openreview.net/forum?id=7sASqAmGaO  (2024)

\bibitem{agem}
Chaudhry, A., Ranzato, M., Rohrbach, M., Elhoseiny, M.: Efficient lifelong learning with a-{GEM}. In: International Conference on Learning Representations (2019), \url{https://openreview.net/forum?id=Hkf2_sC5FX}

\bibitem{simclr}
Chen, T., Kornblith, S., Norouzi, M., Hinton, G.: A simple framework for contrastive learning of visual representations. In: International conference on machine learning. pp. 1597--1607. PMLR (2020)

\bibitem{simsiam}
Chen, X., He, K.: Exploring simple siamese representation learning. In: Proceedings of the IEEE/CVF conference on computer vision and pattern recognition. pp. 15750--15758 (2021)

\bibitem{cheng2023contrastive}
Cheng, H., Wen, H., Zhang, X., Qiu, H., Wang, L., Li, H.: Contrastive continuity on augmentation stability rehearsal for continual self-supervised learning. In: Proceedings of the IEEE/CVF International Conference on Computer Vision. pp. 5707--5717 (2023)

\bibitem{survey_CL}
De~Lange, M., Aljundi, R., Masana, M., Parisot, S., Jia, X., Leonardis, A., Slabaugh, G., Tuytelaars, T.: Continual learning: A comparative study on how to defy forgetting in classification tasks. arXiv preprint arXiv:1909.08383  \textbf{2}(6), ~2 (2019)

\bibitem{w-mse}
Ermolov, A., Siarohin, A., Sangineto, E., Sebe, N.: Whitening for self-supervised representation learning. In: International Conference on Machine Learning. pp. 3015--3024. PMLR (2021)

\bibitem{cassle}
Fini, E., Da~Costa, V.G.T., Alameda-Pineda, X., Ricci, E., Alahari, K., Mairal, J.: Self-supervised models are continual learners. In: Proceedings of the IEEE/CVF Conference on Computer Vision and Pattern Recognition. pp. 9621--9630 (2022)

\bibitem{byol}
Grill, J.B., Strub, F., Altch{\'e}, F., Tallec, C., Richemond, P., Buchatskaya, E., Doersch, C., Avila~Pires, B., Guo, Z., Gheshlaghi~Azar, M., et~al.: Bootstrap your own latent-a new approach to self-supervised learning. Advances in neural information processing systems  \textbf{33},  21271--21284 (2020)

\bibitem{gui2023survey}
Gui, J., Chen, T., Zhang, J., Cao, Q., Sun, Z., Luo, H., Tao, D.: A survey on self-supervised learning: Algorithms, applications, and future trends (2023)

\bibitem{memory1}
Guo, Y., Liu, M., Yang, T., Rosing, T.: Improved schemes for episodic memory-based lifelong learning. In: Advances in Neural Information Processing Systems. vol.~33, pp. 1023--1035. Curran Associates, Inc. (2020), \url{https://proceedings.neurips.cc/paper/2020/file/0b5e29aa1acf8bdc5d8935d7036fa4f5-Paper.pdf}

\bibitem{moco}
He, K., Fan, H., Wu, Y., Xie, S., Girshick, R.: Momentum contrast for unsupervised visual representation learning. In: Proceedings of the IEEE/CVF conference on computer vision and pattern recognition. pp. 9729--9738 (2020)

\bibitem{he2016deep}
He, K., Zhang, X., Ren, S., Sun, J.: Deep residual learning for image recognition. In: Proceedings of the IEEE conference on computer vision and pattern recognition. pp. 770--778 (2016)

\bibitem{hu2021well}
Hu, D., Yan, S., Lu, Q., Hong, L., Hu, H., Zhang, Y., Li, Z., Wang, X., Feng, J.: How well does self-supervised pre-training perform with streaming data? arXiv preprint arXiv:2104.12081  (2021)

\bibitem{huang2023resolving}
Huang, B., Chen, Z., Zhou, P., Chen, J., Wu, Z.: Resolving task confusion in dynamic expansion architectures for class incremental learning. In: Proceedings of the AAAI Conference on Artificial Intelligence. vol.~37, pp. 908--916 (2023)

\bibitem{kalantidis2020hard}
Kalantidis, Y., Sariyildiz, M.B., Pion, N., Weinzaepfel, P., Larlus, D.: Hard negative mixing for contrastive learning. Advances in Neural Information Processing Systems  \textbf{33},  21798--21809 (2020)

\bibitem{kim2022theoretical}
Kim, G., Xiao, C., Konishi, T., Ke, Z., Liu, B.: A theoretical study on solving continual learning. Advances in Neural Information Processing Systems  \textbf{35},  5065--5079 (2022)

\bibitem{kim2020mixco}
Kim, S., Lee, G., Bae, S., Yun, S.Y.: Mixco: Mix-up contrastive learning for visual representation. arXiv preprint arXiv:2010.06300  (2020)

\bibitem{EWC}
Kirkpatrick, J., Pascanu, R., Rabinowitz, N., Veness, J., Desjardins, G., Rusu, A.A., Milan, K., Quan, J., Ramalho, T., Grabska-Barwinska, A., et~al.: Overcoming catastrophic forgetting in neural networks. Proceedings of the national academy of sciences  \textbf{114}(13),  3521--3526 (2017)

\bibitem{krizhevsky2009learning}
Krizhevsky, A., Hinton, G., et~al.: Learning multiple layers of features from tiny images  (2009)

\bibitem{le2015tiny}
Le, Y., Yang, X.: Tiny imagenet visual recognition challenge. CS 231N  \textbf{7}(7), ~3 (2015)

\bibitem{lee2020mix}
Lee, K., Zhu, Y., Sohn, K., Li, C.L., Shin, J., Lee, H.: i-mix: A domain-agnostic strategy for contrastive representation learning. arXiv preprint arXiv:2010.08887  (2020)

\bibitem{li2020compositional}
Li, Y., Zhao, L., Church, K., Elhoseiny, M.: Compositional language continual learning  (2020)

\bibitem{lwf}
Li, Z., Hoiem, D.: Learning without forgetting. IEEE transactions on pattern analysis and machine intelligence  \textbf{40}(12),  2935--2947 (2017)

\bibitem{lin2023speciality}
Lin, Y., Tan, L., Lin, H., Zheng, Z., Pi, R., Zhang, J., Diao, S., Wang, H., Zhao, H., Yao, Y., et~al.: Speciality vs generality: An empirical study on catastrophic forgetting in fine-tuning foundation models. arXiv preprint arXiv:2309.06256  (2023)

\bibitem{lin2022continual}
Lin, Z., Wang, Y., Lin, H.: Continual contrastive learning for image classification. In: 2022 IEEE International Conference on Multimedia and Expo (ICME). pp.~1--6. IEEE (2022)

\bibitem{rgo}
Liu, H., Liu, H.: Continual learning with recursive gradient optimization. In: International Conference on Learning Representations (2022), \url{https://openreview.net/forum?id=7YDLgf9_zgm}

\bibitem{gem}
Lopez-Paz, D., Ranzato, M.A.: Gradient episodic memory for continual learning. In: Guyon, I., Luxburg, U.V., Bengio, S., Wallach, H., Fergus, R., Vishwanathan, S., Garnett, R. (eds.) Advances in Neural Information Processing Systems. vol.~30. Curran Associates, Inc. (2017), \url{https://proceedings.neurips.cc/paper/2017/file/f87522788a2be2d171666752f97ddebb-Paper.pdf}

\bibitem{lump}
Madaan, D., Yoon, J., Li, Y., Liu, Y., Hwang, S.J.: Representational continuity for unsupervised continual learning. In: International Conference on Learning Representations (2022), \url{https://openreview.net/forum?id=9Hrka5PA7LW}

\bibitem{reg2}
Mallya, A., Lazebnik, S.: Packnet: Adding multiple tasks to a single network by iterative pruning. In: 2018 {IEEE} Conference on Computer Vision and Pattern Recognition, {CVPR} 2018, Salt Lake City, UT, USA, June 18-22, 2018. pp. 7765--7773. Computer Vision Foundation / {IEEE} Computer Society (2018). \doi{10.1109/CVPR.2018.00810}, \url{http://openaccess.thecvf.com/content\_cvpr\_2018/html/Mallya\_PackNet\_Adding\_Multiple\_CVPR\_2018\_paper.html}

\bibitem{infomax}
Ozsoy, S., Hamdan, S., Arik, S., Yuret, D., Erdogan, A.: Self-supervised learning with an information maximization criterion. Advances in Neural Information Processing Systems  \textbf{35},  35240--35253 (2022)

\bibitem{er}
Robins, A.: Catastrophic forgetting, rehearsal and pseudorehearsal. Connection Science  \textbf{7}(2),  123--146 (1995)

\bibitem{expanse1}
Rusu, A.A., Rabinowitz, N.C., Desjardins, G., Soyer, H., Kirkpatrick, J., Kavukcuoglu, K., Pascanu, R., Hadsell, R.: Progressive neural networks. CoRR  \textbf{abs/1606.04671} (2016), \url{http://arxiv.org/abs/1606.04671}

\bibitem{gpm}
Saha, G., Garg, I., Roy, K.: Gradient projection memory for continual learning. In: International Conference on Learning Representations (2021), \url{https://openreview.net/forum?id=3AOj0RCNC2}

\bibitem{expanse3}
Sarwar, S.S., Ankit, A., Roy, K.: Incremental learning in deep convolutional neural networks using partial network sharing. IEEE Access  \textbf{8},  4615--4628 (2020). \doi{10.1109/ACCESS.2019.2963056}

\bibitem{reg1}
Serra, J., Suris, D., Miron, M., Karatzoglou, A.: Overcoming catastrophic forgetting with hard attention to the task. In: Dy, J., Krause, A. (eds.) Proceedings of the 35th International Conference on Machine Learning. Proceedings of Machine Learning Research, vol.~80, pp. 4548--4557. PMLR (10--15 Jul 2018), \url{https://proceedings.mlr.press/v80/serra18a.html}

\bibitem{gen_rep}
Shin, H., Lee, J.K., Kim, J., Kim, J.: Continual learning with deep generative replay. In: Guyon, I., Luxburg, U.V., Bengio, S., Wallach, H., Fergus, R., Vishwanathan, S., Garnett, R. (eds.) Advances in Neural Information Processing Systems. vol.~30. Curran Associates, Inc. (2017), \url{https://proceedings.neurips.cc/paper_files/paper/2017/file/0efbe98067c6c73dba1250d2beaa81f9-Paper.pdf}

\bibitem{skean2024frossl}
Skean, O., Dhakal, A., Jacobs, N., Giraldo, L.G.S.: Fro{SSL}: Frobenius norm minimization for self-supervised learning (2024), \url{https://openreview.net/forum?id=1mOeklnLf4}

\bibitem{tang2024kaizen}
Tang, C.I., Qendro, L., Spathis, D., Kawsar, F., Mascolo, C., Mathur, A.: Kaizen: Practical self-supervised continual learning with continual fine-tuning. In: Proceedings of the IEEE/CVF Winter Conference on Applications of Computer Vision. pp. 2841--2850 (2024)

\bibitem{wang2020understanding}
Wang, T., Isola, P.: Understanding contrastive representation learning through alignment and uniformity on the hypersphere. In: International Conference on Machine Learning. pp. 9929--9939. PMLR (2020)

\bibitem{expanse4}
Yoon, J., Kim, S., Yang, E., Hwang, S.J.: Scalable and order-robust continual learning with additive parameter decomposition. In: International Conference on Learning Representations (2020), \url{https://openreview.net/forum?id=r1gdj2EKPB}

\bibitem{expanse2}
Yoon, J., Yang, E., Lee, J., Hwang, S.J.: Lifelong learning with dynamically expandable networks. In: International Conference on Learning Representations (2018), \url{https://openreview.net/forum?id=Sk7KsfW0-}

\bibitem{barlowtwins}
Zbontar, J., Jing, L., Misra, I., LeCun, Y., Deny, S.: Barlow twins: Self-supervised learning via redundancy reduction. In: International Conference on Machine Learning. pp. 12310--12320. PMLR (2021)

\bibitem{zeng2019sese}
Zeng, Z., Xulei, Y., Qiyun, Y., Meng, Y., Le, Z.: Sese-net: Self-supervised deep learning for segmentation. Pattern Recognition Letters  \textbf{128},  23--29 (2019)

\bibitem{zhang2022claire}
Zhang, H., Mueller, F.: Claire: Enabling continual learning for real-time autonomous driving with a dual-head architecture. In: 2022 IEEE 25th International Symposium On Real-Time Distributed Computing (ISORC). pp. 1--10. IEEE (2022)

\bibitem{zhang2017mixup}
Zhang, H., Cisse, M., Dauphin, Y.N., Lopez-Paz, D.: mixup: Beyond empirical risk minimization. In: International Conference on Learning Representations (2018), \url{https://openreview.net/forum?id=r1Ddp1-Rb}

\end{thebibliography}
\clearpage
\appendix
\section{Related Works: Continued}
\label{relatedworks}
\textbf{Self Supervised Learning:} 
SSL divides into two primary sub-areas: 1- Generative self-supervised learning, in which models predict missing data components (e.g., inpainting) \cite{gui2023survey}, and 2- Discriminative self-supervised learning, where models learn representations by distinguishing between different views of identical data (e.g., contrastive learning)\cite{gui2023survey}. As the focus of this work is discriminative SSL, we will offer comprehensive literature on this aspect and utilize SSL to specifically denote discriminative SSL throughout this work.

Recent works in SSL are predominantly categorized into three families: sample contrastive, asymmetric network and dimension-contrastive \cite{skean2024frossl}. 
Sample-contrastive techniques, exemplified by SimCLR \cite{simclr}, MoCO \cite{moco}, SwAV \cite{swav}, and FroSSL \cite{skean2024frossl}, treat to different views of a sample (i.e. different augmented versions) as positive samples, and any other sample as negative. Then, a contrastive loss is employed to bring positive samples closer while pushing negative samples apart. 
Asymmetric-network approaches include SimSiam \cite{simsiam}, BYOL \cite{byol}, and DINO \cite{dino}. These methods require distinct network architectures for input views. While one network serves as the primary network for final use, the other can adopt a different encoder structure, or stop-gradient techniques can be employed within the same architecture as the primary network \cite{byol}.
Dimension-contrastive methods, such as Barlow-Twins \cite{barlowtwins}, VicReg \cite{bardes2022vicreg}, W-MSE \cite{w-mse}, CorInfoMax \cite{infomax}, and FroSSL \cite{skean2024frossl}, focus on reducing redundancy within embedding dimensions.
Such approaches may eliminate the need for negative samples and the requirement for an asymmetric network structure.\\

\noindent \textbf{Continual Learning:} Continual Learning refers to Continual Supervised Learning generally. The main problem of Continual learning is catastrophic forgetting, and recent methods developed various strategies to solve that problem\cite{survey_CL}. These methods can be broadly classified into three distinct approaches:

\begin{enumerate}
    \item \textbf{Rehearsal-based Approaches:} These methods mitigate forgetting by replaying data from previous tasks, either stored in a limited memory \cite{gem, agem, er, memory1} or synthesized by generative models \cite{gen_rep}. Notably, \cite{lwf} employs an experience replay technique, utilizing a stored model from a preceding task as a 'teacher' for knowledge distillation on features pre-classification.

    \item \textbf{Expansion-based Approaches:} To prevent forgetting, these methods dynamically increase the network's capacity in response to new tasks \cite{expanse1,expanse2,expanse3,expanse4}. The most crucial disadvantage of expansion-based methods is large model sizes when the new tasks appear continuously. 

    \item \textbf{Regularization-based Approaches:} This category encompasses methods that adjust the optimization process by introducing task-specific regularization. Such regularization aims to align the optimal parameters of new tasks with those of prior tasks, thereby minimizing forgetting \cite{EWC,rgo, reg1, reg2}. A notable variation within this approach is the Gradient Projection Memory (GPM) technique \cite{gpm}, which adjusts model gradients on a per-layer basis to ensure updates are orthogonal to the gradient subspace of preceding tasks, thus preserving prior learning. Federated Orthogonal training (FOT) \cite{bakman2024federated} is another work mainly proposed for distributed learning settings that modify the global updates of new tasks so that they are orthogonal to previous tasks’ activation principal subspace.
\end{enumerate}

Although these methods listed above show impressive results on Continual Supervised Learning, they are not that effective in CSSL \cite{cassle,sycon}.\\

\section{Task Confusion Experiments}
\label{task_cobfusion_hps}
%
In Section \ref{motivation}, we stated our hypothesis, which is as follows,
\begin{adjustwidth}{0.9cm}{0.9cm}
  {\textit{The task confusion problem in Contrastive SSL methods arises primarily from the inability to train the model with different classes belonging to different tasks concurrently.}}
\end{adjustwidth}
Here, we perform a data-incremental learning-based ablation study to further analyze our hypothesis.\\

\begin{figure}[h]
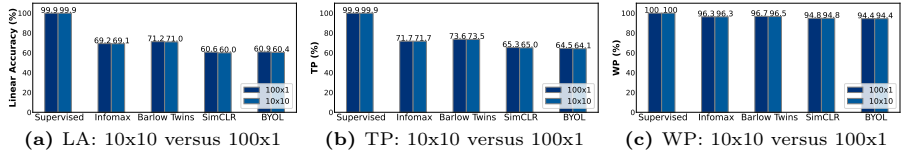

\centering
\begin{subfigure}{.32\textwidth}
    \centering
    \includegraphics[width=.98\linewidth]{figures/DIL_results/CIL_train_lin_di.pdf}  
    \caption{LA: 10x10 versus 100x1}
    \label{SUBFIGURE LABEL 1}
\end{subfigure}
\begin{subfigure}{.32\textwidth}
    \centering
    \includegraphics[width=.98\linewidth]{figures/DIL_results/CIL_train_tp_di.pdf}  
    \caption{TP: 10x10 versus 100x1}
    \label{SUBFIGURE LABEL 2}
\end{subfigure}
\begin{subfigure}{.32\textwidth}
    \centering
    \includegraphics[width=.98\linewidth]{figures/DIL_results/CIL_train_wp_di.pdf}  
    \caption{WP: 10x10 versus 100x1}
    \label{}
\end{subfigure}
\caption{Training LA, WP, and TP performance of contrastive SSL methods and supervised learning on the CIFAR100 Dataset for both 100x1 and 10x10 DIL-minibatch settings. Figures (a), (b), and (c) demonstrate that the 10x10 setting performs as good as the 100x1 setting in terms of LA, TP, and WP, respectively. }
\label{fig:dil_bw_results}
\end{figure}

\noindent \textbf{Data-Incremental Learning: An Ablation Study}\\
 In this ablation study, we aim to analyze that the performance drop in LA and TP observed in Table \ref{fig:bw_results} was because of the class split across tasks, not because of the data split across tasks. For that, we explore both self-supervised and supervised learning in another fairly simple but representative data-incremental setup, where each task has all classes data but data is split across tasks. To be more specific, our experiment setup is as follows: we follow a data-incremental setup with a sequence of tasks \(\mathcal{T}_1, \mathcal{T}_2, ..., \mathcal{T}_T\) that have same set of classes. We consider the CIFAR100 dataset and split the 50,000 data in 10 tasks with each task containing all 100 classes data and total of 5,000 samples per task. Further, we assume that tasks change after each iteration, i.e., mini-batches are sampled from different tasks at each iteration. To remove the forgetting effect, we assume that tasks can be revisited, i.e., task 2 follows task 1, task 3 follows task 2, and so on. The repeatability of tasks ensures that the SSL learner does not forget the previous knowledge. For simplicity, we refer to this experimental setup, 10x10 data-incremental learning across mini-batches, 10x10 DIL-minibatch because the data is divided into 10 tasks where each task contains 5000 samples. Further, to show how the performance of the methods changes in DIL-minibatch setting, we also compare it to the regular setting where we sample uniformly random from the whole training data. We call this regular training setting as 100x1 DIL-minibatches because there is 1 task containing all 100 classes and all the data. Here again, we report the training accuracy of the methods because we only care about the methods' capability of creating linearly separable features on the data they trained on. The training accuracy of these methods is reported in Figure \ref{fig:dil_bw_results}.
 If data split across batches/tasks would have been the reason for the accuracy drop, we would observe an accuracy drop in these settings as well. However, in contrast to what we observed in Figure \ref{fig:bw_results}, we did not observe any accuracy drop in average linear accuracy between 10x10 and 100x1 DIL experiment settings for both supervised as well as self-supervised learning settings. This reaffirms our hypothesis that the task confusion problem in Contrastive SSL methods arises primarily from the inability to train the model with different classes belonging to different tasks concurrently. 

\section{Continual Learning Experiments}
\label{further_experiments}
In this section, we provide further evaluations: an ablation study to compare different design components of CroMo-Mixup, buffer size versus accuracy performance, the model's generalization performance, and catastrophic forgetting analysis of CroMo-Mixup with the progression of time. 
\subsection{An Ablation Study: Different Design Components' Performance}
In Table \ref{task_ablation}, we evaluate three key components of our design model, cross-task data mixup, cross-model feature mixup, and distillation $(\zeta)$ on CIFAR100-Split5 with Barlow-Twins. First, we compare the cross-task data mixup without adding any other component. For input mixup comparison, we have two choices:  within-task data mixup and cross-task data mixup. We find that the cross-task data mixup enhances the performance by 8\% accuracy gain as compared to the within-task setting, as shown in the first two rows of Table \ref{task_ablation}. Next, we compare output mixup, again we have two cases, the same model feature mixup that uses only the current model to obtain embeddings, and cross-model feature mixup, the proposed formulation. We note that the cross-model feature mixup enhances the performance by 1.6\% accuracy gain as compared to the same-model mixup for the cross-task data mixup. This highlights the significance of the proposed CroMo-Mixup formulation. Additionally, we also notice that distillation improves the performance by another 1.5\% accuracy gain with these components. Hence, cross-task mixup, cross-model mixup, and distillation components help in achieving the highest performance among other design choices of these components.

\begin{table}[h]
\centering
\caption{An Ablation Study on Different Design Components of CroMo-Mixup}
\fontsize{6}{8}\selectfont
\begin{tabular}{|c|c|c|c|}
\hline
\multicolumn{1}{|c|}{\textbf{Input Mixup}} & \multicolumn{1}{c|}{\textbf{Output Mixup}} & \multicolumn{1}{c|}{\textbf{Distillation} $(\zeta)$} & \multicolumn{1}{c|}{\textbf{Accuracy(\%)}} \\
\hline
within-task & same-model & 0 & 54.16 \\
cross-task & same-model & 0 & 62.31 \\
\hline
cross-task & same-model & 0 & 62.31 \\
cross-task & cross-model & 0 & 63.94 \\
\hline
cross-task & same-model & 1 & 64.35 \\
cross-task & cross-model & 1 & 65.48 \\
\hline
\end{tabular}
\label{task_ablation}
\end{table}


\subsection{Buffer Size versus Accuracy Performance}
\begin{figure}
    \centering
    \includegraphics[width=.5\linewidth]{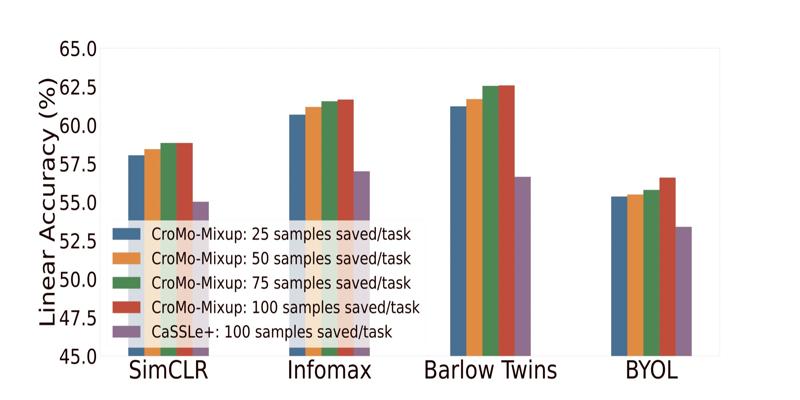}
    \caption{Average linear accuracy performance of CIFAR100-Split10 CroMo-Mixup with varying buffer size options: 25, 50, 75, 100 samples/task. As buffer size reduces, the model performance decreases from 1-2 \% at most; however, it still outperforms the SOTA baseline CaSSLe+ for each respective SSL baseline.}
    \label{fig:mem}
\end{figure}

To analyze the impact of buffer size on the performance of CroMo-Mixup, we experiment with four buffer size options, 25, 50, 75, and 100 samples saved per task for CIFAR100-Split10. As shown in Figure \ref{fig:mem}, we observe that with smaller buffer sizes, model performance drops within the 1-2 percentage. However, even in the smaller budget of 25 samples/task, CroMo-Mixup outperforms the SOTA baseline CaSSLe+, that saves 100 samples per task, for each respective SSL baseline. 


\subsection{Out-of-Distribution Performance}
To evaluate the model's generalization performance, we test the ResNet-18 model trained with the CIFAR100-Split5 setting on two other datasets, CIFAR10 and Oxford Flower102. We compare the model trained on five different settings; offline, FineTune, Ering, CaSSLe+, and CroMo-Mixup. For all these baselines, at the end of the training, we freeze the encoder and train a linear classifier on the training dataset of CIFAR10 and Flower102 for 200 epochs. We use an SGD optimizer with a learning rate of 0.2 for the downstream task. We evaluate its performance on the test set of each respective dataset. We present the results in Table \ref{ood}. The key observation from these experiments is that CroMo-Mixup outperforms CaSSL+ on both datasets. This highlights the significance of CroMo-Mixup to generalize better to the unseen distributions. 
\begin{table}[h!]
\centering
\caption{Average Linear Accuracy Performance evaluation of CIFAR100-Split5 trained model on the test data of CIFAR10 and Flower 102 datasets}
\fontsize{8}{10}\selectfont
\begin{tabular}{|c|c|c|c|c|c|}
\hline
\multicolumn{1}{|c|}{\textbf{Dataset}} & \multicolumn{1}{|c|}{\textbf{Offine}} & \multicolumn{1}{|c|}{\textbf{FineTune}} & \multicolumn{1}{|c|}{\textbf{Ering}} & \multicolumn{1}{|c|}{\textbf{CaSSLe+}} & \multicolumn{1}{|c|}{\textbf{CroMo-Mixup}} \\
\hline
CIFAR10 & 82.27 & 71.71 & 75.08 & 77.16 & 80.36\\
Flower102 & 51.94 & 34.05 & 38.00 & 44.17 & 49.03\\
\hline
\end{tabular}
\label{ood}
\end{table}


\subsection{Catastrophic Forgetting Mitigation Performance}
\begin{figure}[h!]
\centering
\begin{subfigure}{.32\textwidth}
    \centering
    \includegraphics[width=.98\linewidth]{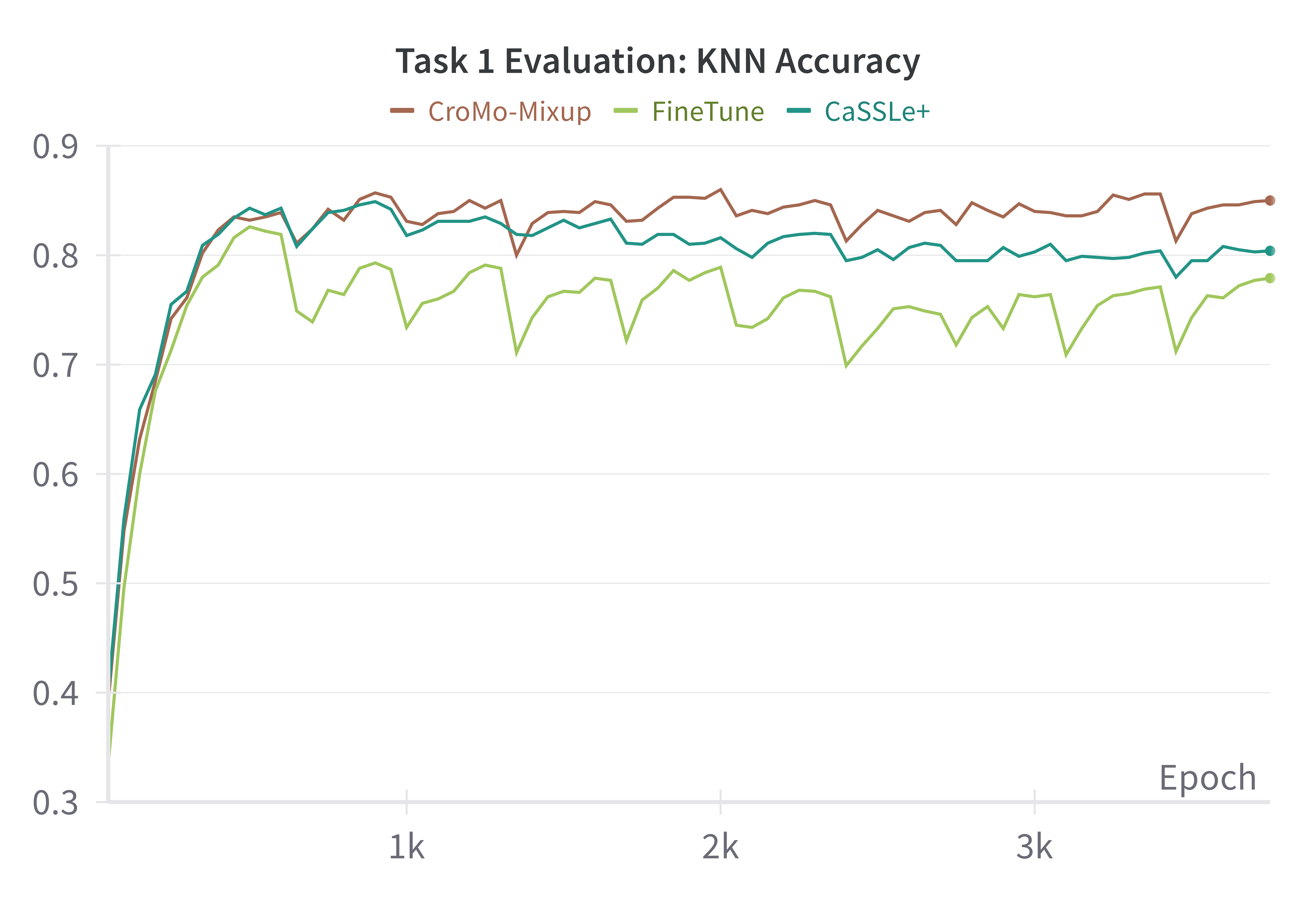}  
    \caption{Task 1 Evaluation}
    \label{task1}
\end{subfigure}
\begin{subfigure}{.32\textwidth}
    \centering
    \includegraphics[width=.98\linewidth]{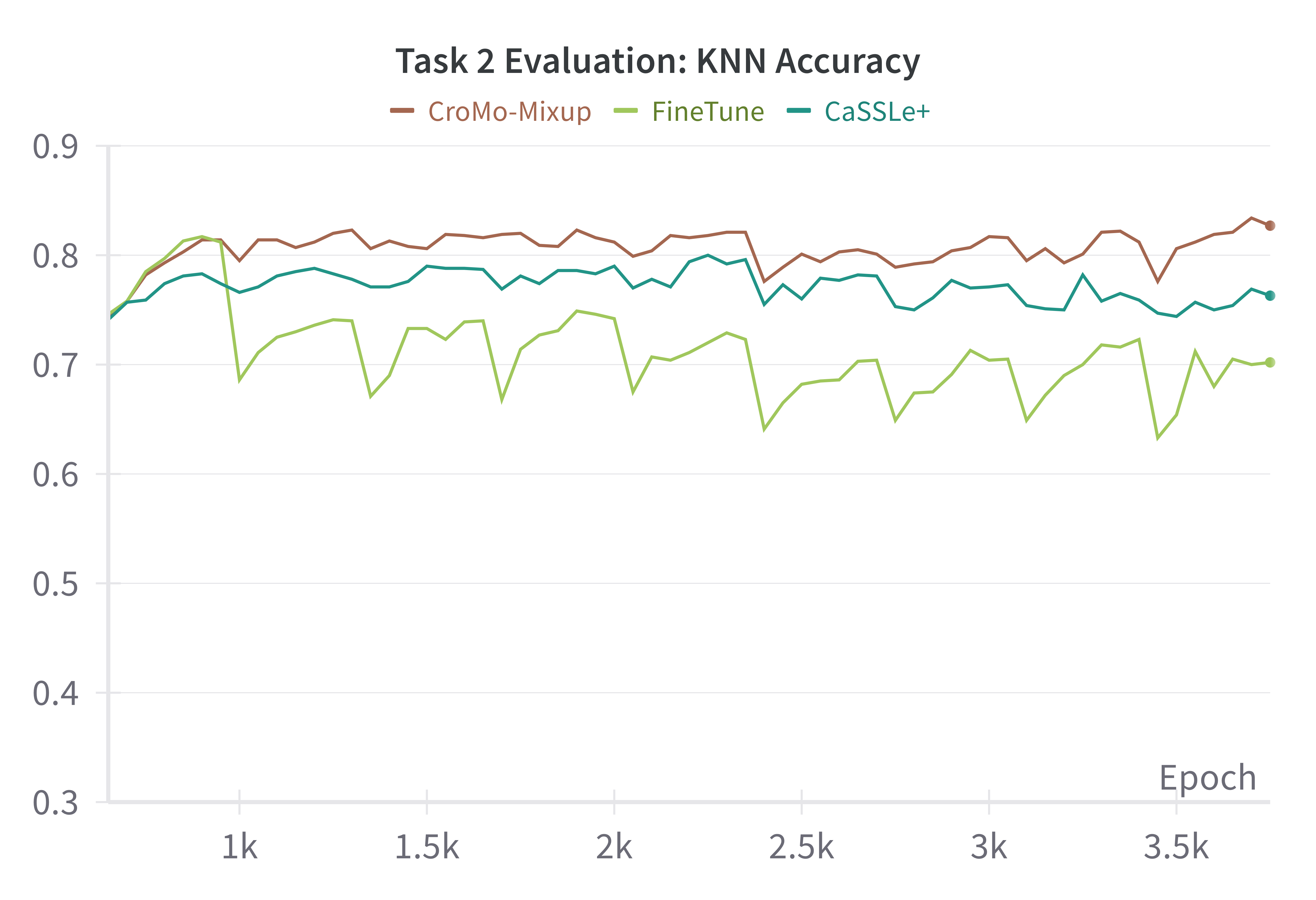}  
    \caption{Task 2 Evaluation}
    \label{task2}
\end{subfigure}
\begin{subfigure}{.32\textwidth}
    \centering
    \includegraphics[width=.98\linewidth]{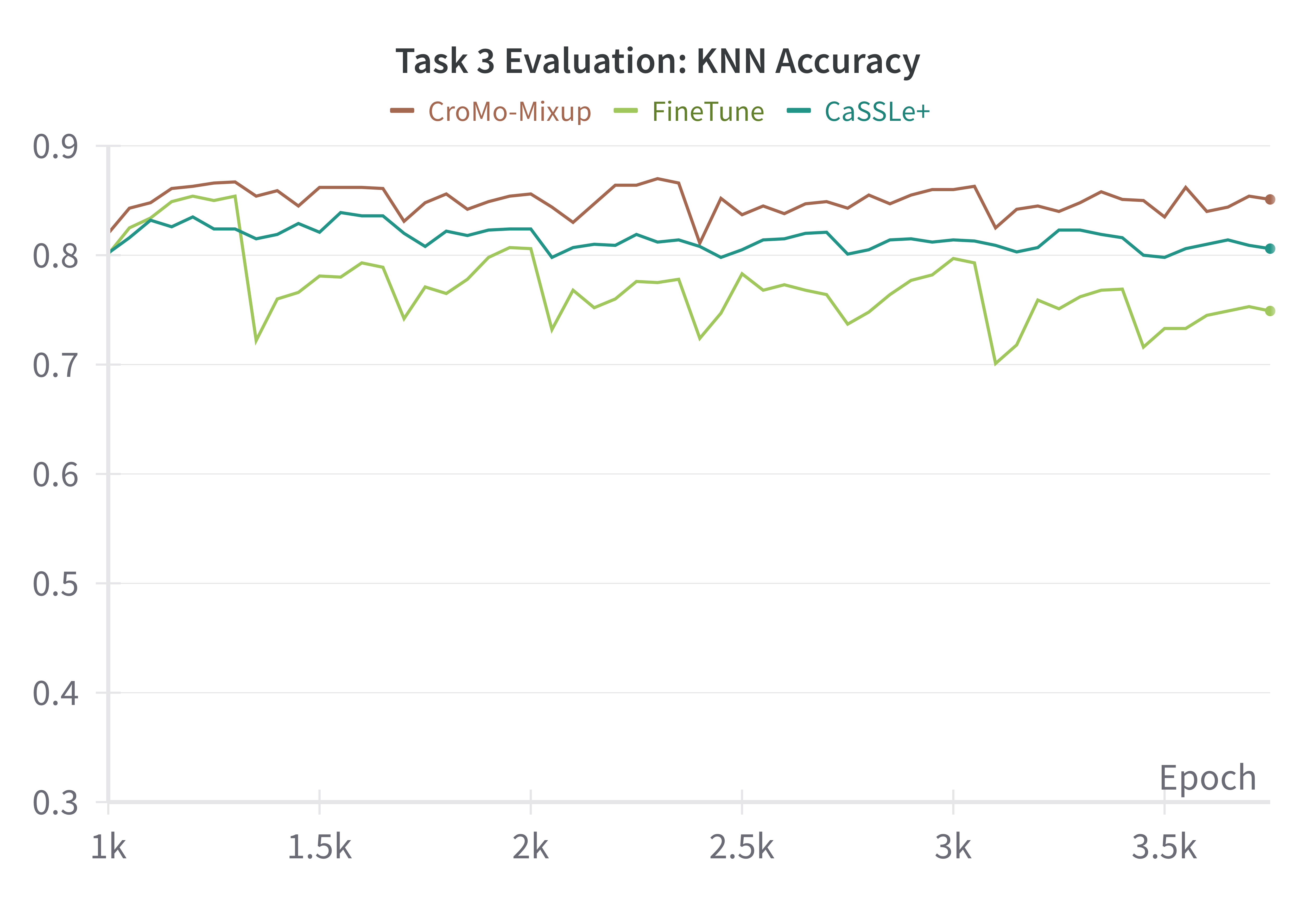}  
    \caption{Task 3 Evaluation}
    \label{task3}
\end{subfigure}
\caption{KNN Accuracy evaluation of Task 1, 2, and 3 on their respective test sets for CIFAR100-Split10 with the Barlow Twins SSL baseline. }
\label{fig:catastraphic_forgetting}
\end{figure}

We analyze CIFAR100-Split10 as an example to compare the performance of CaSSLe+ and CroMo-Mixup to address catastrophic forgetting. Figure \ref{fig:catastraphic_forgetting}, shows the K-nearest neighbors (KNN) accuracy of the model with the progression of tasks. We use (K=200) to report the KNN accuracy.

\begin{figure}[h!]
\centering
\begin{subfigure}{.32\textwidth}
    \centering
    \includegraphics[width=.98\linewidth]{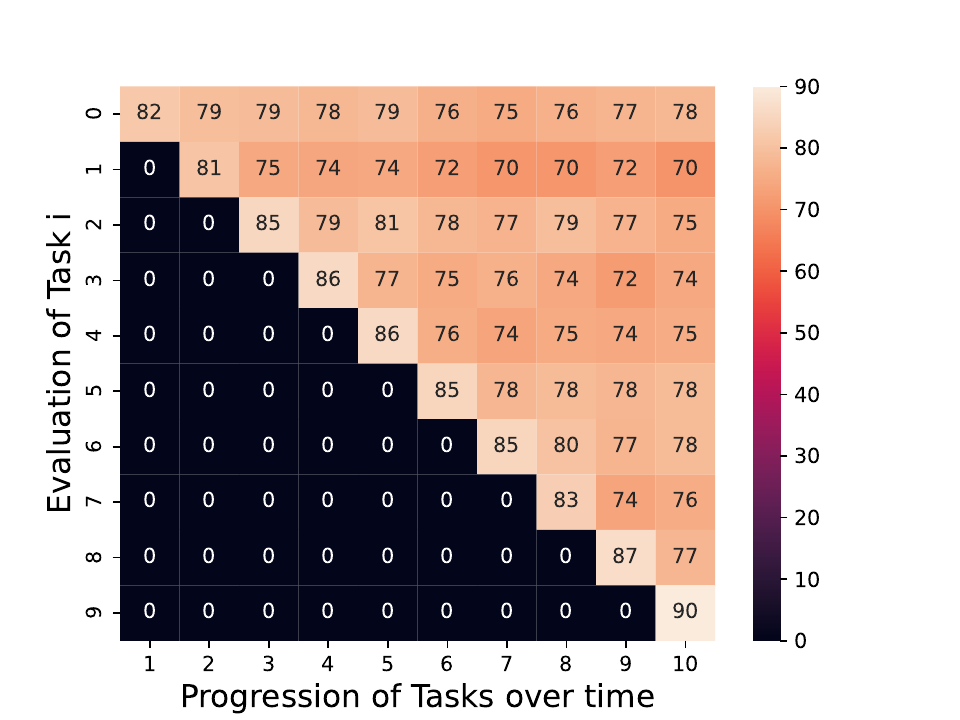}  
    \caption{FineTune}
    \label{tc1}
\end{subfigure}
\begin{subfigure}{.32\textwidth}
    \centering
    \includegraphics[width=.98\linewidth]{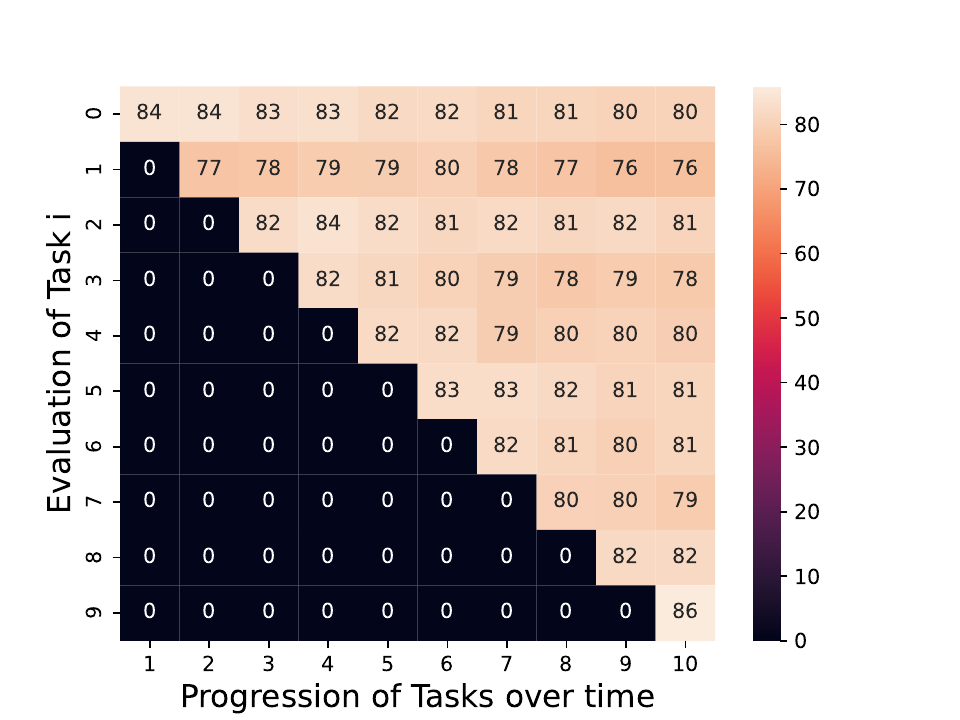}  
    \caption{CaSSLe+}
    \label{tc2}
\end{subfigure}
\begin{subfigure}{.32\textwidth}
    \centering
    \includegraphics[width=.98\linewidth]{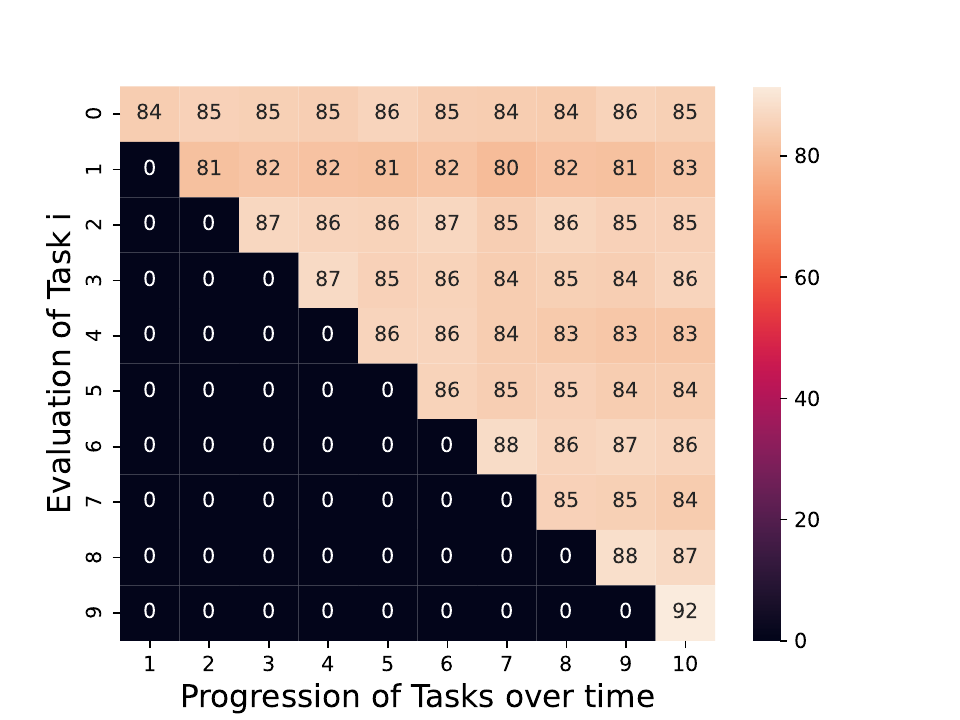}  
    \caption{CroMo-Mixup}
    \label{tc3}
\end{subfigure}
\caption{KNN Accuracy Performance of different tasks for CIFAR100-Split10 with the Barlow-Twins baseline. 0 here indicates the time instances when that task was not available and, therefore, is not evaluated. Comparing the last column of each task, CroMo-Mixup either achieves equal or higher accuracy than the SOTA baseline, CaSSLe+.}
\label{fig:task_confusion_matrix}
\end{figure}

From Figure \ref{fig:catastraphic_forgetting}, we can see that finetune baseline model struggles with catastrophic forgetting. We observe a significant accuracy drop as the model learns new tasks. Though CaSSLe+ and CroMo-Mixup help to alleviate catastrophic forgetting, CroMo-Mixup outperforms CaSSLe+ by achieving 4.5\%, 6\%, and 4.5\% higher KNN test accuracy at the end of CL training as shown in Figures \ref{task1}, \ref{task2} and \ref{task3}. For the remaining tasks, we present the confusion matrix for FineTune, CaSSLe+, and CroMo-Mixup in Figures \ref{tc1}, \ref{tc2}, and \ref{tc3}, respectively. First we note that all these matrices are diagonal dominant. Further, for the last column of each task, CroMo-Mixup either achieves equal or higher performance than CaSSLe+, which highlights the model's ability to alleviate forgetting. 

\section{Hyper-Parameter Configurations}
In this section, we provide the hyper-parameter configurations and the SSL loss function descriptions.
\label{cssl_hps}
\subsection{Hyper-Parameter Configurations in CSSL Experiments}
Table \ref{tab:details_cssl} presents the important hyperparameters for all SSL methods. Most hyperparameters are selected from the original code bases of these works. For CSSL hyper-parameters such as epochs/task, learning rate scaling, etc, optimal hyperparameters are found by doing a hyperparameter grid search on basic fine-tuning (FT). Subsequently, the identified optimal set of hyperparameters is uniformly applied across all CSSL methods for consistency in evaluation. Furthermore, the buffer size details are provided in Table \ref{experiment_setting}. For sampling data from the memory buffer, we use a batch size of 64, which we selected by hyperparameter tuning on ER.

\begin{table}

\caption{Hyperparameter Settings for the CSSL experiments}
\label{tab:details_cssl}
\centering
\smallskip\noindent
\resizebox{.98\linewidth}{!}{
\begin{tabular}{c||ccccc}
\hline
\begin{tabular}[c]{@{}c@{}}CIFAR-10 / CIFAR-100\\  / Tiny-Imagenet\end{tabular} & \multicolumn{1}{c|}{\begin{tabular}[c]{@{}c@{}}Cor-Infomax\end{tabular}} & \multicolumn{1}{c|}{\begin{tabular}[c]{@{}c@{}}SimCLR \end{tabular}} & \multicolumn{1}{c|}{\begin{tabular}[c]{@{}c@{}}BYOL\end{tabular}} & \multicolumn{1}{c}{\begin{tabular}[c]{@{}c@{}}Barlow-Twins\end{tabular}} \\  \hline
\begin{tabular}[c]{@{}c@{}}Batch Size\end{tabular}                     & \multicolumn{1}{c|}{512/512/256}                                                       & \multicolumn{1}{c|}{512/512/256}                                                         & \multicolumn{1}{c|}{256}                                             & \multicolumn{1}{c}{256}                                              \\ \hline
Learning rate                                                                   & \multicolumn{1}{c|}{0.1/0.1/0.5}                                                       & \multicolumn{1}{c|}{0.6/0.6/0.3}                                                         & \multicolumn{1}{c|}{1.0 / 1.0 / 0.3}                                             & \multicolumn{1}{c}{0.3}                                              \\ \hline
Optimizer                                                                       & \multicolumn{1}{c|}{SGD}                                                       & \multicolumn{1}{c|}{SGD}                                                         & \multicolumn{1}{c|}{LARS}                                             & \multicolumn{1}{c}{LARS}                                                                                                                                                                                                                                                                                                                       \\ \hline
Weight decay                                                                    & \multicolumn{1}{c|}{1e-4}                                                      & \multicolumn{1}{c|}{5e-4}                                                        & \multicolumn{1}{c|}{1e-5}                                                        & \multicolumn{1}{c}{1e-4}                                                        \\ \hline
\begin{tabular}[c]{@{}c@{}}Projection layer\\  (dim)\end{tabular}               & \multicolumn{1}{c|}{64/128/64}                                                      & \multicolumn{1}{c|}{128/128/2048}                                                        & \multicolumn{1}{c|}{4096}                                                        & \multicolumn{1}{c}{2048}                                                        \\ \hline
\begin{tabular}[c]{@{}c@{}}Prediction layer\\  (dim, for BYOL)\end{tabular}     & \multicolumn{1}{c|}{-}                                                         & \multicolumn{1}{c|}{-}                                                           & \multicolumn{1}{c|}{4096}                                                           & \multicolumn{1}{c}{-}                                        &       \\ \hline
\begin{tabular}[c]{@{}c@{}}Temperature\\  ($\tau$)\end{tabular}                 & \multicolumn{1}{c|}{-}                                                       & \multicolumn{1}{c|}{0.5}                                                         & \multicolumn{1}{c|}{-}                                                           & \multicolumn{1}{c}{-}                                                            \\ \hline
\end{tabular}
}
\end{table}
 
\label{HPs}

\begin{table}
\centering
\caption{CSSL Experiment Setup Details}
\fontsize{8}{10}\selectfont
\begin{tabular}{|c|c|c|c|c|c|}
\hline
\multicolumn{1}{|c|}{Experiment} & \multicolumn{1}{|c|}{Total \# } & \multicolumn{1}{|c|}{Classes} & \multicolumn{1}{|c|}{Total \# } & \multicolumn{1}{|c|}{Samples} & \multicolumn{1}{|c|}{Total \# of } \\
Name & of Classes & per Task & of Tasks & saved/Task & Samples per Task \\
\hline
\hline
cifar10-Split2 & 10 & 5 & 2 & 500 & 25,000\\
cifar100-Split5 & 100 & 20 & 5 & 500 & 10,000 \\
cifar100-Split10 & 100 & 10 & 10 & 100 & 5,000\\
tinyImageNet-Split10 & 200 & 20 & 10 & 100 &50,000 \\
\hline
\end{tabular}
\label{experiment_setting}
\end{table}


\subsection{Hyper-Parameter Configurations in Task Confusion Experiments}
For the task confusion experiments, we use ResNet-18 encoder to train on the CIFAR100 dataset. We select the hyper-parameters following the original papers of the respective works for 100x1 and use the same settings for 10x10 experiments. Details of some of the hyper-parameters are provided in Table \ref{cifar100_hps}. 

\begin{table}
\centering
\caption{Hyper-Parameter Settings for the Task Confusion Experiments }
\fontsize{8}{10}\selectfont
\begin{tabular}{|c||c|c|c|c|c|}
\hline
\multicolumn{1}{|c||}{\textbf{Name}}  & \multicolumn{1}{|c|}{\textbf{CorInfomax}} & \multicolumn{1}{|c|}{\textbf{Barlow-Twins}}  & \multicolumn{1}{|c|}{\textbf{SimCLR}} &\multicolumn{1}{|c|}{\textbf{BYOL}} & \multicolumn{1}{|c|}{\textbf{Supervised}} \\
\hline
Optimizer& SGD &LARS&LARS&LARS&SGD\\
\hline
 lr &0.5&0.3& 0.6&1.0 &0.075\\
\hline
epochs &1000&1000&1000&1000&200\\
\hline
batch size &512&256&512&256&128\\
\hline
\end{tabular}
\label{cifar100_hps}
\end{table}

\subsection{SSL Loss functions}
Here, we provide the details of the loss functions of the four SSL baselines on which we deployed CroMo-Mixup.
\subsubsection{BYOL Loss Function}
BYOL employs a momentum encoder, where gradients are backpropagated only through the first augmentation of the data, and the second augmentation encoder network is updated by an exponential moving average (EMA). It employs an MSE-based loss function, which essentially enforces consistency between the $l_2$ normalized embedding vectors of $z^{1}$ and $z^{2}$ as $|| q^{1} - z^{2}||_{2}^{2}$, where $q^{1} = h(z^{1})$ and $h(.)$ is the predictor head.
\subsubsection{SimCLR Loss Function}
SimCLR uses InfoNCE loss function which treats the second of augmentation of an image as its positive, and every other image in the mini-batch as negative. The InfoNCE loss function is calculated on the feature embeddings as follows,
\begin{equation}
    \mathcal{L}_{\text{InfoNCE}} = -\text{log}\frac{\text{exp}(z_{i}^{1}, z_{i}^{2})}{\sum_{z_{j}\in \eta(i)} \text{exp}(z^{1}_{i}, z_{j})}
\end{equation}
where $\eta(i)$ contains all the negative samples of image indexed at $i$ and all the embedding vectors are $l_2$ normalized.
\subsubsection{Barlow-Twins Loss Function}
Barlow-twins employs a cross-correlation based loss function, as shown below,
\begin{equation}
    \mathcal{L}_{\text{BT}} = \sum_{i} (1-\mathcal{C}_{ii})^2 + \lambda \sum_{i} \sum_{i \neq j} \mathcal{C}^{2}_{ij}
    \label{BT}
\end{equation}
where $\lambda$ is a positive hyper-parameter. The first term is invariance term that makes the feature embeddings invariant to the data augmentation, whereas second term is redundancy reduction term which reduces the redundancy in the output units. The cross-correlation $\mathcal{C}$ is computed between the feature embeddings of first and second augmentation along the batch dimension as given below,
\begin{equation}
    \mathcal{C}_{ij} = \frac{\sum_{b} z^{1}_{b,i}z^{2}_{b,j}}{\sqrt{\sum_{b}({z^{1}_{b,i})^{2}}} \sqrt{\sum_{b}({z^{2}_{b,j})^{2}}}}
\end{equation}
where $b$ signifies the index for the batch samples and $i$, $j$ identify the index of the vector dimension of the output embeddings.
\subsubsection{CorInfomax Loss Function}
CorInfomax uses log determinant mutual information (LDMI) criterion for self-supervised learning. Its objective is essentially an estimate of LDMI between the two augmented views of model output embeddings vectors as given below,
\begin{equation}
    \mathcal{L}_{\text{CorInfomax}} = -\text{log det} (\hat{R}_{z}^{(1)}[l] + \epsilon I) -\text{log det} (\hat{R}_{z}^{(2)}[l] + \epsilon I) + 2\frac{2}{\epsilon N}||Z^{(1)}[l] - Z^{(2)}[l]||^{2}_{F}
    \label{infomax}
\end{equation}
where $l$ identifies the batch number, $||.||_{F}$ is Frobenius norm. Further, $R_{z}^{(2)}[l]$ and $R_{z}^{(2)}[l]$ are auto-covariance estimates calculated as follows,
\begin{equation}
    \hat{R}_{z}^{(1)}[l] = \lambda \hat{R}^{(1)}_{z}[l-1]
    (1-\lambda) \frac{1}{N} \bar{Z}^{(1)}[l]\bar{Z}^{(1)}[l]^{T}
\end{equation}
Likewise, $\hat{R}_{z}^{(2)}[l]$ is estimated based on the current and last batch statistics. Note that $\bar{Z}^{(2)}[l]$ are the mean-centralized feature embeddings, $\bar{Z}^{(2)}[l] = Z^{(2)}[l] - \mu^{(2)}[l]I^{T}_{N}$ where $\mu^{(2)}$ is the mean estimate of the current and old batches. 


\label{ssl_loss}

\end{document}